\documentclass[11pt]{article}
\pdfoutput=1

\usepackage[preprint]{acl}

\usepackage{times}
\usepackage{latexsym}

\usepackage[T1]{fontenc}

\usepackage[utf8]{inputenc}

\usepackage{microtype}

\usepackage{inconsolata}

\usepackage{graphicx}
\usepackage{xcolor}
\usepackage{listings}

\usepackage{CJKutf8}
\usepackage[utf8]{inputenc}
\usepackage[T1]{fontenc}
\newcommand{\ja}[1]{\begin{CJK}{UTF8}{ipxm}#1\end{CJK}}

\newcommand{\numlingfeatures}{45~}
\newcommand{\numstgyfeatures}{11~}

\usepackage{booktabs}
\usepackage{tabularx}
\usepackage{threeparttable}
\usepackage{multirow}
\usepackage{array}

\usepackage{amsmath}
\usepackage{amssymb}
\usepackage{amsfonts}

\usepackage{longtable}
\usepackage{tablefootnote}

\title{Who Laughs with Whom? Disentangling Influential Factors in Humor Preferences across User Clusters and LLMs}

\author{
    Soichiro Murakami$^{1}$, \ Hidetaka Kamigaito$^{1,2}$, Hiroya Takamura$^3$, \ Manabu Okumura$^3$ \\
  $^1$CyberAgent, $^2$Nara Institute of Science and Technology, $^3$Institute of Science Tokyo \\
  {\tt murakami\_soichiro@cyberagent.co.jp}, \\
  {\tt kamigaito.h@is.naist.jp}, {\tt \{takamura,oku\}@pi.titech.ac.jp} \\
}

\begin{document}
\maketitle
\begin{abstract}
Humor preferences vary widely across individuals and cultures, complicating the evaluation of humor using large language models (LLMs).
In this study, we model heterogeneity in humor preferences in Oogiri, a Japanese creative response game, by clustering users with voting logs and estimating cluster-specific weights over interpretable preference factors using Bradley-Terry-Luce models.
We elicit preference judgments from LLMs by prompting them to select the funnier response and found that user clusters exhibit distinct preference patterns and that the LLM results can resemble those of particular clusters.
Finally, we demonstrate that, by persona prompting, LLM preferences can be directed toward a specific cluster.
The scripts for data collection and analysis are publicly available to support reproducibility.
\end{abstract}

\section{Introduction\label{section:introduction}}
Large language models (LLMs) have garnered significant attention, as they are capable of creative reasoning akin to humans.
Humor understanding and generation, which require contextual and nuanced understanding, provide a useful testbed for evaluating the creative capabilities.
However, because humor is highly subjective and exhibits cultural dependence, the quantitative measurement and replication of funniness have long been a challenging problem in natural language processing~\cite{Loakman2025INLG}.
In this study, we focus on Oogiri, a Japanese creative response game.
In Oogiri, which is characterized by a question-answer style of humor, participants produce witty responses to a given prompt.

\begin{figure}[t]
  \centering
  \includegraphics[width=1\linewidth]{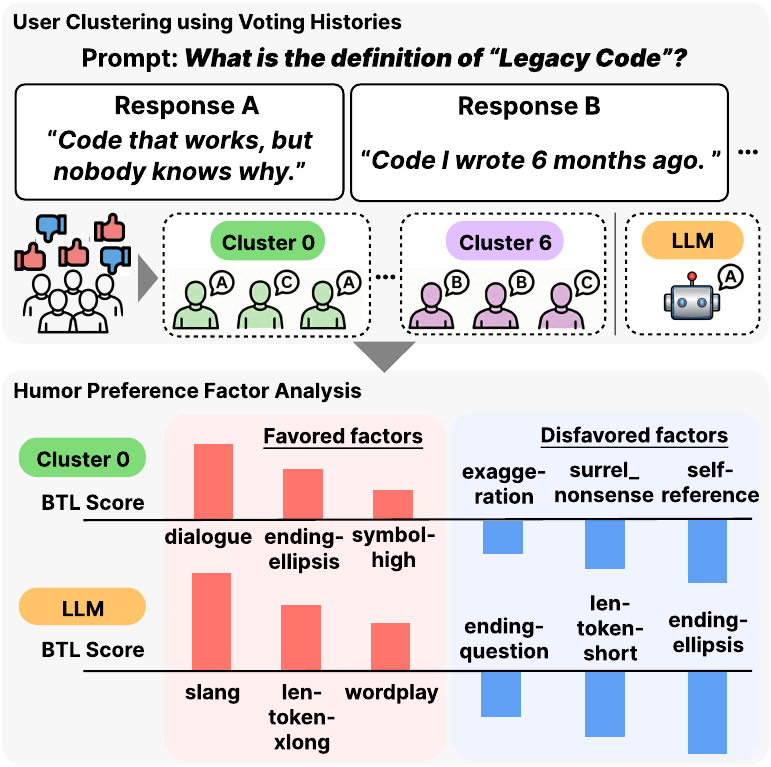}
  \caption{Overview of humor-preference factor analysis across user clusters and LLMs. Users are clustered based on their voting history, and humor-preference factors are analyzed using the Bradley-Terry-Luce model.}
  \label{fig:overview}
\end{figure}

Improving humor understanding and response generation requires both clarifying human humor preferences and identifying the gaps between LLMs and humans. In this study, we analyze the factors underlying human humor preferences and examine the gaps in humor preferences between LLMs and humans, which have already been investigated in several studies, while two challenges still remain.
First, previous analyses of human humor preferences do not consider individual preferences~\cite{murakami2025oogiricorpus,sakabe2025assessingcapabilitiesllmshumora}.
Previous work typically collected ratings from multiple human annotators for each humorous text and aggregated them into an ``overall score'' by averaging.
However, in subjective evaluation tasks, including humor judgments, inter-annotator agreement is often low~\cite{celikyilmaz2021evaluationtextgenerationsurvey}. Thus, it is reasonable to assume that humor preferences differ across users~\cite{Chakrabarty2019RBMJokeRecommender,Zhang2024HumorAI}.
Given the subjectivity of humor and its cultural dependence, humor research should go beyond aggregated preferences and account for individual preferences~\cite{Cabitza_Campagner_Basile_2023}.

Second, the gaps between LLM and individual human humor preferences remain still unclear.
\citet{sakabe2025assessingcapabilitiesllmshumora} analyzed the differences between human and LLM humor preferences, showing that humans prioritize \textit{empathy}, whereas LLMs prioritize \textit{novelty}.
However, because their analysis relied on the ``aggregated humor preferences'' obtained from multiple user ratings and compared only LLMs with the overall user population, it neither accounted for individual user preferences nor examined the alignment between individual human and LLM preferences.
Therefore, they remain still unclear whether the LLM humor preferences are intrinsically different from those of humans and whether they align with those of certain users. 

To address these issues, we analyze the factors affecting humor preferences at the user cluster level, using the voting data collected from the Oogiri platform (Figure~\ref{fig:overview}).
Because each vote is associated with a user ID, we represent each user with a voting history-based vector and cluster these vectors to identify the groups with similar preferences.
For each cluster, we fit a Bradley--Terry--Luce (BTL) model \cite{bradley1952rank} to estimate the weights of preference factors, including linguistic features (e.g., length) and humor-strategy labels (e.g., black joke).
To enable a direct comparison with LLMs' preferences, we additionally construct LLM preference data by asking the model to select the funniest response for each prompt.
The BTL model is then fitted to the LLM data using the same set of factors, and the resulting weights are compared with those of each user cluster.
This analysis quantifies the preference gaps between LLMs and user clusters, and tests whether any clusters exhibit preference patterns aligned with those of LLMs.

Based on this user cluster-aware analysis, we investigate the following research questions:
\begin{itemize}
  \item RQ1: What kinds of humor do different user clusters prefer?
  \item RQ2: Do LLMs align with the preferences of any specific user clusters?
  \item RQ3: Can we align LLM preferences with those of a specific user cluster?
\end{itemize}
In answering these questions, our contributions are four-fold.
First, we quantified the variation in humor preferences across user clusters by estimating cluster-specific preference factors.
Second, we characterized differences between LLM and human humor preferences in aggregate and identified cases in which the LLM results resemble particular clusters in terms of factor weights.
Third, we demonstrated that persona prompting can increase the alignment between an LLM and a specific user cluster.
Finally, we release our data collection and analysis scripts to support reproducibility.\footnote{\url{https://github.com/CyberAgentAILab/oogiri-dataset-builder}}
We hope that our findings will provide useful insights for advancing humor understanding and generation that accounts for heterogeneous humor preferences.

\section{Related Work\label{section:related_work}}
\looseness=-1
With the rapid progress in LLMs, humor understanding and generation have received renewed attention~\cite{amin-burghardt-2020-survey}.
In natural language processing, humor has been studied in tasks such as humor recognition~\cite{mihalcea-strapparava-2005-making,Miller2017SemEvalPuns,weller-seppi-2019-humor}, humor generation~\cite{he-etal-2019-pun,luo-etal-2019-pun}, and, more recently, the evaluation of LLMs' humor capabilities~\cite{jentzsch-kersting-2023-chatgpt,hessel-etal-2023-androids}.
Humor spans diverse forms, such as puns and irony, and corresponding datasets and methods have been developed~\cite{ritchie-2005-computational,hossain-etal-2022-memosen,hessel-etal-2023-androids}.
This study focuses on Oogiri, a traditional Japanese humor format.
Oogiri is a question-answer style of humor, whereby a witty response is produced for a given prompt.
Recently, online Oogiri platforms have emerged. \citet{Zhong_2024_CVPR} collected data from Bokete\footnote{\url{https://bokete.jp}} to build the Oogiri-GO dataset, thereby accelerating research in this area~\cite{sakabe2025assessingcapabilitiesllmshumora,murakami2025oogiricorpus}.

\looseness=-1
Oogiri is related to ``Caption This'' (e.g., The New Yorker's Cartoon Caption Contest), where a participant creates a funny caption for a given panel or photo~\cite{hessel-etal-2023-androids,tanaka-etal-2024-content,Zhang2024HumorAI}.
Although both \textit{Caption This} and Oogiri require the generation of a witty response conditioned on a prompt, Oogiri in this work is text-based, whereas \textit{Caption This} is multimodal.
However, \citet{zhou-etal-2025-bridging} reported that state-of-the-art LLMs often struggle with visually grounded humor, which can undermine the reliability of LLM-based preference estimation in multimodal settings.
Accordingly, we focused on text-based Oogiri, deferring multimodal settings to future studies.

Quantifying the factors that shape humor preferences can help advance research on humor understanding~\cite{murakami2025oogiricorpus}.
In particular, contrasting human and LLM preferences provides a clearer picture of current LLM humor understanding and potential improvements.
Recently, \citet{sakabe2025assessingcapabilitiesllmshumora} analyzed these differences and found that humans prefer \textit{empathy}, whereas LLMs prefer \textit{novelty}; however, their analysis was based on humor preferences aggregated across multiple users and thus did not address how well LLM preferences align with those of individual users.
In a related study, \citet{Chakrabarty2019RBMJokeRecommender} constructed user clusters from joke voting data for recommender systems but did not analyze cluster-specific preference patterns in detail.
Therefore, we analyze humor-preference factors across user clusters to characterize the humor preferences of LLMs and the cluster-level differences.

\section{Construction of Analytical Dataset\label{section:dataset_construction}}

Existing Oogiri datasets consist of prompt-response pairs and vote counts, indicating the funniness of each response; however, they lack information on which user voted for which response, making it impossible to track individual humor preferences.
Therefore, we extended the existing Japanese Oogiri dataset, the Oogiri-Corpus~\cite{murakami2025oogiricorpus}, to construct a new dataset that includes user-level voting data for each response.

\paragraph{Source Dataset}
The Oogiri-Corpus is a prompt-response pair dataset collected from the Oogiri platform, Oogiri Sogo,\footnote{\url{https://chinsukoustudy.com}} comprising 908 prompts and 82,536 responses, each associated with a vote count.
Users can cast up to three votes per prompt on this platform.
The three votes can be distributed flexibly; for example, a user may allocate two votes to one response and one vote to another.
Because the user IDs were attached to the voting data, the voting histories of individual users can be tracked.

\paragraph{Dataset Construction Process}
We built the analytical dataset with user-level voting data in two steps: (1) web-crawling the votes, including user IDs for all prompt-response pairs from the source site, and (2) filtering.
To ensure reliability, active users with at least 100 total votes were first selected, ensuring that their humor preferences are well represented.
Only the votes from these active users were retained, recomputing each response's vote count.
Responses with fewer than three votes were then removed to maintain high response quality.

\paragraph{Dataset Statistics}
The dataset contained 908 prompts, 14,389 responses, and 57,751 votes from 276 users (35.6 users per prompt on average).

\section{Method}
Figure~\ref{fig:overview} summarizes our analysis pipeline.
To account for heterogeneity in humor preferences, we analyzed preference factors at the user-cluster level.
We first constructed user representations and clustered users based on their voting histories (\S\ref{section:user_clustering}), then defined humor preference factors for each prompt-response pair (\S\ref{section:humor_preference_factors}), and finally estimated factor weights with an BTL model (\S\ref{sec:preference_modeling}).

\subsection{User Representation and Clustering\label{section:user_clustering}}
The users were clustered based on their voting histories.
Each user $u$ is represented by a sparse voting history vector $\mathbf{x}_u \in \mathbb{R}^N$, where $N$ is the number of responses in the dataset and $\mathbf{x}_u[i]$ is the number of votes cast by $u$ for response $i$ ($0$ if never selected). 
To reduce the influence of responses frequently chosen by many users, we applied term frequency-inverse document frequency (TF-IDF) reweighting to obtain $\tilde{\mathbf{x}}_u$.
Subsequently, to mitigate sparsity, we computed a 100-dimensional representation by applying a truncated singular value decomposition (SVD), $\mathbf{z}_u = \mathrm{SVD}_{100}(\tilde{\mathbf{x}}_u)$, and normalized it as $\mathbf{y}_u = \mathbf{z}_u / \|\mathbf{z}_u\|_2$ to control for scale differences across users.
Finally, we clustered $\{\mathbf{y}_u\}$ using K-means clustering.
We report the clustering configuration and robustness analyses in Appendix~\ref{appendix:clustering_robustness}.

\subsection{Humor Preference Factors\label{section:humor_preference_factors}}
\begin{table*}[t]
\centering
\small
\begin{threeparttable}
\begin{tabularx}{\linewidth}{l r X}
  \toprule
  \multicolumn{1}{c}{\textbf{Group}} & 
  \multicolumn{1}{c}{\textbf{\#}} & 
  \multicolumn{1}{c}{\textbf{Brief definition}} \\
  \midrule
  Linguistic Features  & & \\
  \hspace{1em}Basic           & 11 & Surface statistics of each response (e.g., character count, character type ratio).  \\
  \hspace{1em}Morphological   & 10 & Morphological analysis features of each response (e.g., part-of-speech ratio,  word count).  \\
  \hspace{1em}Special symbols & 5 & Usage of special symbols in each response (e.g., quotation marks, parentheses, slang phrases).  \\
  \hspace{1em}Sentence-ending & 9 & Sentence-ending patterns of each response (e.g., whether it ends with a symbol (?, !)).  \\
  \hspace{1em}Writing style   & 4 & Japanese writing-style indicators in each response (e.g., polite/casual forms, exaggeration). \\
  \hspace{1em}Relational & 6 & Features capturing relations between a prompt and a response. (e.g., length ratio)  \\
  \midrule
  \multirow{2}{*}{Humor strategy} & \multirow{2}{*}{\numstgyfeatures} & Interpretable multilabel annotations derived from humor theories to capture nuanced humor beyond linguistic features (e.g., \texttt{incongruity}, \texttt{black\_joke\_satire}, \texttt{self\_reference}.)  \\
  \bottomrule
\end{tabularx}
\end{threeparttable}
\caption{Summary of humor preference factors used in our BTL analysis. A full list of features and their definitions is provided in Appendix \ref{appendix:definition_of_humor_preference_factors}.}
\label{table:humor-preference-factors-summary}
\end{table*}

We designed two main types of features that influence user humor preferences.
Table \ref{table:humor-preference-factors-summary} summarizes the humor preference factors used in this study.
The first group consists of linguistic features, which are basic features extracted directly from the prompts and responses (\S\ref{section:linguistic_features}). We defined  \numlingfeatures linguistic features.
The second group comprises humor strategy types that capture more nuanced aspects of humor (\S\ref{section:humor_strategy_labels}).
Specifically, we assigned \numstgyfeatures different humor strategy labels to each prompt-response pair, covering various humor strategies such as ``black joke'' and ``parody.''
We defined these features with reference to prior research on humor factor analysis  \cite{murakami2025oogiricorpus} and humor theories \cite{Morreall2024PhilosophyHumor}.
An overview of each feature group is provided below, with detailed definitions available in Appendix \ref{appendix:definition_of_humor_preference_factors}.

\subsubsection{Linguistic Features\label{section:linguistic_features}}
We designed \numlingfeatures linguistic features, ranging from basic features, such as text length and part-of-speech (POS) ratios, to features based on the relationship between the prompt and response.
These features were categorized into six subgroups: basic, morphological analysis, sentence-ending pattern, special symbol, writing-style, and prompt-response relational features.
The first five subgroups are based on the linguistic characteristics of the responses, whereas the last subgroup is based on the relationship between the prompt and response.

\subsubsection{Humor Strategy Labels\label{section:humor_strategy_labels}}

To capture more nuanced aspects of humor that cannot be represented by linguistic features alone, we annotated each prompt-response pair with humor strategy labels.
We defined the \numstgyfeatures strategy labels based on humor theories \cite{Morreall2024PhilosophyHumor} and prior research~\cite{murakami2025oogiricorpus}.
Theories of humor, which aim to explain the essence of humor and the mechanisms by which humans perceive humor, have been studied in fields such as psychology, sociology, and linguistics.
For example, \texttt{incongruity} represents unexpected twists and surprising connections, aligning with \textit{incongruity theory}~\cite{mcdonald2013philosophy}, whereas \texttt{black\_joke\_satire} encompasses dark humor and social commentary, relating to \textit{benign violation theory}~\cite{McGraw2010-kg}.

\paragraph{Annotation Process}
\looseness=-1
We annotated all prompt responses using GPT-5.1.
We carefully designed the system prompts, including detailed definitions and annotation examples for each label.
To ensure reliability, we followed a self-consistency protocol~\cite{wang-etal-2023-self-consistency}, conducting three trials per response and determining the final labels by majority voting.
Each response can be assigned multiple labels.
We additionally validated the labels via a crowdsourced study (1,100 instances, five annotators per instance; 81.4\% judged appropriate overall, with an average agreement rate of 76.2\%); see Appendix~\ref{appendix:humor_strategy_labeling} for details and examples.

\subsection{Preference Modeling\label{sec:preference_modeling}}
We formalized our approach to analyze humor preference factors by building it on the DecipherPref framework~\cite{hu-etal-2023-decipherpref}.
DecipherPref converts features into categorical and interpretable factors and applies the BTL model~\cite{bradley1952rank,luce1959individual} to analyze the factors that influence pairwise preference judgments in a summarization task (e.g., length, linguistic quality, content accuracy).
In our setting, pairwise preference judgments were modeled as comparisons between the above factors, and each factor's relative strength was estimated using the BTL model.

\looseness=-1
Let $\mathcal{D} = \{(p_i, \mathbf{r}_i, \mathbf{v}_i)\}_{i=1}^{M}$ be a collection of $M$ prompts, where $p_i$ is the $i$-th prompt; $\mathbf{r}_i = (r_i^1, r_i^2, \ldots, r_i^{n_i})$ is the response list for prompt $p_i$; and $\mathbf{v}_i = (v_i^1, v_i^2, \ldots, v_i^{n_i})$ is the vote-count list such that $v_i^j \in \mathbb{N}$ is the vote count for response $r_i^j$; and $n_i$ is the number of responses for prompt $p_i$.

We defined a finite vocabulary $\mathbb{F}$ of $K$ interpretable factors to characterize the prompt-response pairs.
Each factor $f \in \mathbb{F}$ represents a categorical property of a prompt-response pair.
Specifically, we considered two types of factors: (i) linguistic features such as character length and punctuation patterns (\S\ref{section:linguistic_features}) and (ii) humor strategy labels (\S\ref{section:humor_strategy_labels}).
For continuous-valued linguistic features, we discretized the values into quartile-based categorical bins following DecipherPref \cite{hu-etal-2023-decipherpref}; for example, character length is represented as \texttt{len-char-\{short|medium|long|xlong\}} (see Appendix~\ref{appendix:binning_countinous_features} for details).
For any response $r$, let $f(r) \subseteq \mathbb{F}$ be the subset of factors present in $r$.
Factors are binary or multilevel categorical variables, and in the multilevel case, each level is treated as a separate binary factor.

\looseness=-1
We derived pairwise comparisons from vote counts.
For each prompt $p_i$, we compared all pairs of responses $(r_i^j, r_i^k)$ where $j \neq k$; if $v_i^j > v_i^k$, we treated $r_i^j$ as the winner and $r_i^k$ as the loser, and discarded ties ($v_i^j = v_i^k$), as they provide no preference information.
For each winner-loser pair $(r_i^j, r_i^k)$, factor-level comparisons were extracted by defining
\begin{equation*}
F_i^+ = f(r_i^j) \setminus f(r_i^k), \qquad F_i^- = f(r_i^k) \setminus f(r_i^j),
\end{equation*}
where $F_i^+$ and $F_i^-$ denote the factors unique to the winner and loser, respectively.
Factors appearing in both responses were ignored because they do not provide explanatory power for the preference judgment.
Each factor $f \in F_i^+$ was treated as having ``beaten'' every factor $g \in F_i^-$. Thus, one response-level comparison yielded $|F_i^+| \times |F_i^-|$ pairwise outcomes at the factor level.

\looseness=-1
The relative strength of the factors was modeled using the BTL model. %
Each factor $k \in \mathbb{F}$ was associated with a real-valued parameter $\theta_k \in \mathbb{R}$.
For any ordered pair of factors $(k, \ell)$, the probability that factor $k$ beats factor $\ell$ is:
\begin{equation*}
P(k \succ \ell) = \frac{\exp(\theta_k)}{\exp(\theta_k) + \exp(\theta_\ell)}. 
\end{equation*}
This formulation assumes that the probability of one factor beating another depends only on the difference $\theta_k - \theta_\ell$.
The parameters $\hat{\boldsymbol{\theta}}$ were estimated using the Luce spectral ranking (LSR) algorithm~\cite{maystre2015fast}, which is a spectral estimator for models based on Luce's choice axiom~\cite{luce1959individual}.
In our implementation, we used the \texttt{choix} library\footnote{\url{https://choix.lum.li/}} to compute the spectral ranking.
The estimated parameters $\hat{\theta}_k$ provide a ranking of the factors based on their influence on human preferences. Factors with higher $\hat{\theta}_k$ values are strongly associated with preferred responses, whereas factors with lower values are associated with less-preferred responses.
Note that this factor-level model does not capture interaction effects or context-dependent humor, and because some factors co-occur, the estimated scores should not be interpreted as independent causal effects of isolated features (see Appendix~\ref{appendix:factor_correlation} for a correlation analysis among factors).

\section{LLM Preference Data Collection\label{section:collecting_llm_humor_preferences}}
To analyze the gaps between LLM and human preferences for RQ2, we defined a funniest response selection task to collect LLM humor preferences.

\paragraph{Task Definition}
Let $\mathcal{P}$ denote the set of prompts. 
For each prompt $p \in \mathcal{P}$, the LLM is presented with a set of candidate responses $\mathcal{R}(p) = \{r_{p,1}, \ldots, r_{p,K_p}\}$ and instructed to select exactly one response that it finds the funniest.\footnote{For BTL estimation, we converted each selection into a pairwise comparison by treating the selected response as the winner and each non-selected response as the loser.}
The selection is recorded as index $y_p \in \{1, \ldots, K_p\}$ and the chosen response as $\hat r_p = r_{p,y_p}$.
Repeating this procedure over prompts yields a dataset $\{(p, \mathcal{R}(p), y_p)\}_{p \in \mathcal{P}}$, in which each entry consists of a prompt, its associated response set, and the index of the response selected by the LLM as the funniest.\footnote{To address API variability, we called the API three times per prompt with randomly permuted response orders; see Appendix~\ref{appendix:overview_of_persona_prompting}. Self-consistency of these three calls is reported in Appendix~\ref{appendix:llm_self_consistency}.}
This dataset effectively captures the humor preferences of the LLM, enabling us to analyze and compare the LLM's humor preferences with human humor-preference data.

\paragraph{Dataset}
For this task, we constructed an evaluation dataset from the analysis dataset described in \S\ref{section:dataset_construction}.
To avoid unreliable evaluations with insufficient choices, prompts with fewer than five responses were excluded from the analysis dataset.
Consequently, the evaluation dataset comprised 897 unique prompts and 14,352 responses with an average of approximately 16 responses per prompt.

\paragraph{Models}
We analyzed humor-preference variation across three state-of-the-art LLMs: Gemini 3 Pro, GPT-5.1, and Claude Sonnet 4.5.

\paragraph{Persona Prompting}
Prior research has shown that changing the prompts provided to LLMs can affect their response characteristics~\cite{he2024doespromptformattingimpact}.
In this preference collection process, we used persona prompting \cite{tseng-etal-2024-two} as a case study, to investigate how different personas influence humor preferences.
Specifically, we hypothesized that explicitly assigning different personas to LLMs may influence their humor preferences.
If humor preferences vary by persona, we assume that each persona can potentially align with the preferences of specific user clusters.
Thus, we analyzed the relationship between the humor preferences of LLM personas and user clusters.
Specifically, we defined seven personas:
\texttt{\{male,female\}\_20}, \texttt{\{male,female\}\_45}, \texttt{\{male,female\}\_65}, and \texttt{no\_persona}.
For example, the \texttt{male\_20} persona represents a 20-year-old male university student well-versed in trending jokes, memes, and internet vocabulary.
We also included \texttt{no\_persona} as a control condition, whereby no specific persona is provided for the system prompt.
To implement these personas, unique prompts were incorporated for each persona into the system prompt.
We carefully designed these persona prompts to reflect the characteristics and perspectives associated with each persona.
The details of the persona prompts are presented in Appendix~\ref{appendix:overview_of_persona_prompting}.

\section{Experimental Results\label{section:results}}
In the experiments, we analyzed the factors underlying humor preferences for each user cluster and LLM using the BTL model.
This section presents the experimental results that answer the three research questions introduced in \S\ref{section:introduction}.
We first report the results of user clustering based on voting histories (\S\ref{section:results_user_clustering}) and then present the results of the humor preference factor analysis (\S\ref{section:results_of_humor_preference_factor_analysis}).

\subsection{Results of User Clustering\label{section:results_user_clustering}}
\begin{figure}[t]
  \centering
  \includegraphics[width=1\linewidth]{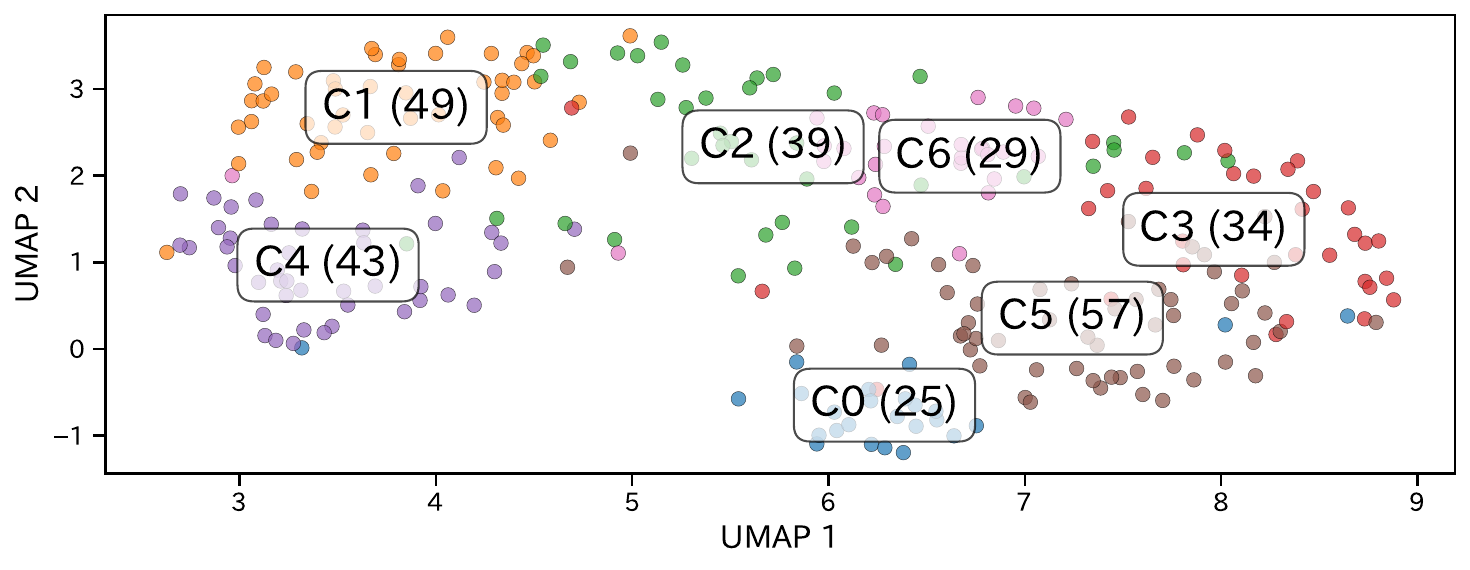}
  \caption{UMAP visualization of user clusters. Different colors represent different clusters. C0 to C6 denote the respective user clusters and the number of users in each cluster is indicated in parentheses.}
  \label{fig:kmeans-user-clusters-umap}
\end{figure}

\looseness=-1
The users were clustered based on their voting histories using K-means clustering.
We identified $K=7$ user clusters: among $K=2$ to $19$, $K=7$ achieved the highest silhouette score~\cite{Rousseeuw1987-silhouette}, and we selected it as a practical trade-off, as larger $K$ values make cluster-level BTL estimation more sparse, whereas smaller $K$ values may merge users with different preferences.
The resulting partition is reasonably stable: it shows moderate agreement across 100 random seeds (median ARI 0.420, NMI 0.555) and with alternative clustering methods (ARI $\approx$ 0.42--0.43), and it is not extremely sensitive to nearby choices of $K$ (see Appendix~\ref{appendix:clustering_robustness} for the full scores and robustness analyses).
In the subsequent analysis (\S\ref{section:results_of_humor_preference_factor_analysis}), we present the results based on these clusters.

For visualization, we embedded the users into two dimensions using uniform manifold approximation and projection (UMAP)~\cite{mcinnes2018umap} (see Figure~\ref{fig:kmeans-user-clusters-umap}).
Although clusters are observable, their boundaries are not sharply separated, consistent with a low silhouette score ($s=0.0258$); we therefore interpret the clusters not as cleanly separated user types but as exploratory groupings for analyzing heterogeneous humor preferences.
Nevertheless, we show that clusters exhibit distinct humor-preference factors in \S\ref{section:humor-preference-factors-user-clusters}.

\subsection{Analysis of Humor Preference Factors\label{section:results_of_humor_preference_factor_analysis}}
\begin{table*}[t]
  \centering
  \small
  \begin{threeparttable}
    \setlength{\tabcolsep}{2.5pt}
		    \begin{tabularx}{\textwidth}{@{}c@{\hspace{1pt}} >{\raggedright\arraybackslash}X >{\raggedright\arraybackslash}X @{}}
      \toprule
      \multicolumn{1}{c}{\textbf{Cluster}} & \multicolumn{1}{c}{\textbf{Preferred factors (Top-3; BTL)}} & \multicolumn{1}{c}{\textbf{Dispreferred factors (Bottom-3; BTL)}} \\
      \midrule
	      \multirow{1}{*}{C0} & parentheses; dialogue; sentences-many & self\_reference; surreal\_nonsense; length-ratio-short  \\
	      \multirow{1}{*}{C1} & self\_reference; ending-adjective; personification & prompt-proper-noun; exaggeration-rule; ending-ellipsis\\
	      \multirow{1}{*}{C2} & self\_reference; mini\_story; ending-question & prompt-proper-noun; ending-ellipsis; exaggeration-rule \\
	      \multirow{1}{*}{C3} & parentheses; ending-ellipsis; space-high & mini\_story; exaggeration; prompt-verb \\
	      \multirow{1}{*}{C4} & ending-ellipsis; self\_reference; parentheses & slang; exaggeration-rule; meta \\
	      \multirow{1}{*}{C5} & slang; exaggeration-rule; prompt-proper-noun & surreal\_nonsense; ending-adjective; len-char-xlong \\
	      \multirow{1}{*}{C6} & surreal\_nonsense; prompt-proper-noun; parody & ending-ellipsis; slang; parentheses \\
      \bottomrule
    \end{tabularx}
    \caption{Cluster-wise preferred and dispreferred humor factors estimated by the BTL model (top-3 and bottom-3). Full results of the cluster-wise BTL scores are reported in Appendix~\ref{appendix:full_btl_scores_for_user_cluster_analysis}.}
    \label{tab:cluster_btl_summary}
  \end{threeparttable}
\end{table*}

\begin{figure*}[t]
  \centering
  \includegraphics[width=1\linewidth]{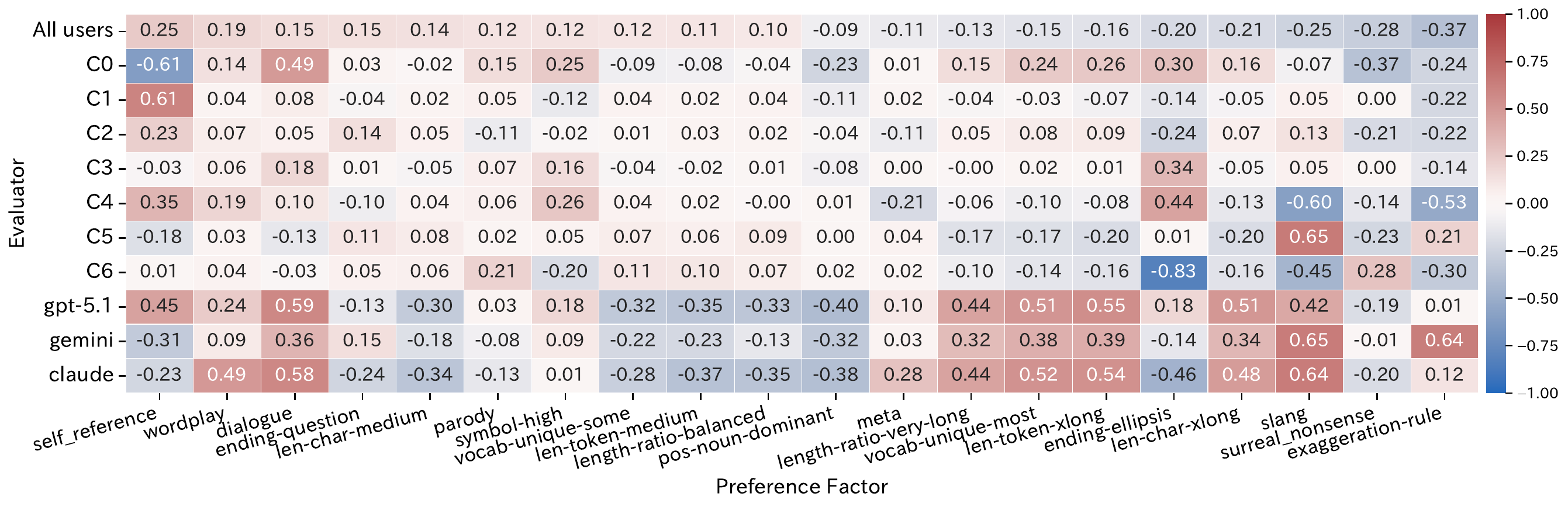}
  \caption{BTL scores of humor preference factors for each user cluster and LLM. C0 to C6 represent each user cluster. ``All users'' indicates the BTL scores calculated using all users without clustering.
  Each LLM shows the BTL scores calculated using humor preference data in the \texttt{no\_persona} setting.
  Due to space limitations, only the top 10 and bottom 10 factors based on the BTL scores of ``All users'' are displayed here. The BTL scores for all factors are provided in Appendices~\ref{appendix:full_btl_scores_for_user_cluster_analysis} and \ref{appendix:full_btl_scores_for_llm_analysis}.}
  \label{fig:all-users-vs-llm-btl-score-heatmap-top10-bottom10}
\end{figure*}

\subsubsection{RQ1: Humor Preference Factors across User Clusters\label{section:humor-preference-factors-user-clusters}}
To answer RQ1, we report the humor preference factors favored by each user cluster.
Table~\ref{tab:cluster_btl_summary} presents the top-3 and bottom-3 factors for each user cluster based on BTL scores, and in Figure~\ref{fig:all-users-vs-llm-btl-score-heatmap-top10-bottom10} the BTL scores are visualized for a broader set of factors as a heatmap (including those for the LLM) to provide a more detailed view.
Our key findings from user cluster analysis are as follows:

\paragraph{User Clusters Exhibit Distinct Preferences }
Each user cluster exhibited a distinct humor preference factor.
For example, as shown in Table~\ref{tab:cluster_btl_summary} and Figure~\ref{fig:all-users-vs-llm-btl-score-heatmap-top10-bottom10}, C0 prefers longer responses that contain dialogue (\texttt{dialogue}) and multiple sentences (\texttt{sentences-many}), whereas C5 tends to dislike overly long responses (\texttt{len-char-xlong}). 
Additionally, factors such as slang and self-deprecating humor show significant variations in preferences across clusters.
For instance, C1 favors self-deprecating humor (\texttt{self\_reference}), whereas C0 tends to dislike it.
This indicates that the influence of humor preference factors varies according to user cluster, reflecting the subjectivity of humor preferences.
To verify that these cluster-level differences do not arise simply from partitioning users into smaller groups, we compared the K-means clusters with 30 random partitions of the same group sizes: the mean factor-wise variance of BTL scores was higher for the K-means clusters (0.0145 vs. 0.0083 $\pm$ 0.0010; one-sided empirical test, $p = 0.032$), suggesting that the clusters capture non-random preference heterogeneity (see Appendix~\ref{appendix:random_partition_baseline}).

\paragraph{Common Preference Factors also Exist across User Clusters}
Some factors exhibited consistent BTL scores across user clusters.
For example, wordplay-based humor (\texttt{wordplay}) and appropriate response length (\texttt{len-char-medium}) show positive scores in most user clusters (Figure~\ref{fig:all-users-vs-llm-btl-score-heatmap-top10-bottom10}).
Similarly, overly long responses (\texttt{len-char-xlong}) show negative scores in many clusters.

\begin{figure}[t]
  \centering
  \includegraphics[width=1\linewidth]{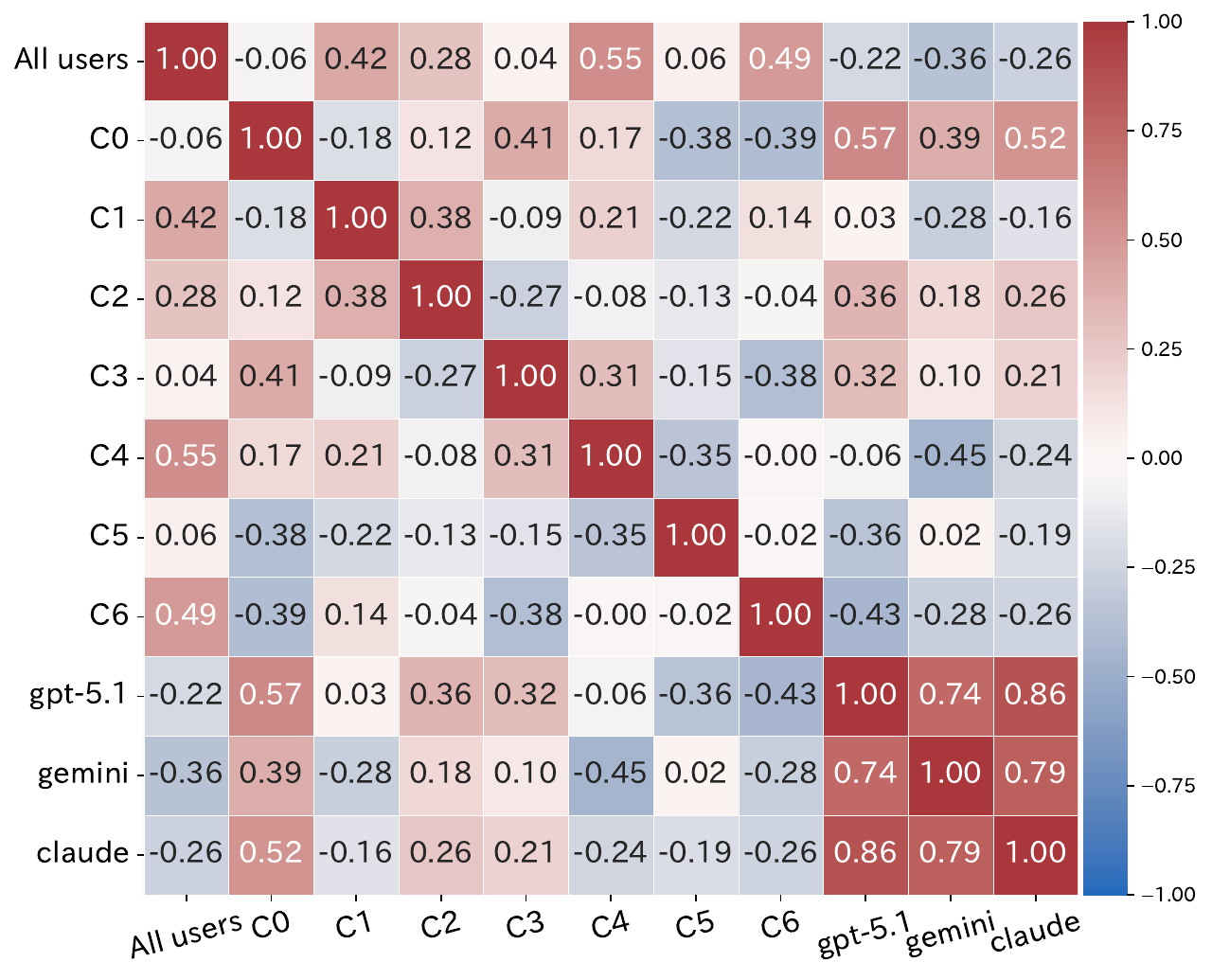}
  \caption{Pearson's correlation matrix of BTL scores between user clusters and LLMs.
   By comparing user clusters and LLMs, we can identify the LLMs that align with the humor preferences of specific user clusters.}
  \label{fig:user-cluster-btl-pearson}
\end{figure}

\paragraph{Correlation Analysis of Humor Preferences between User Clusters}
To quantitatively assess the differences in humor preferences among user clusters, we calculated the Pearson correlation coefficients between the BTL scores of each user cluster.
Figure \ref{fig:user-cluster-btl-pearson} shows the Pearson correlation matrix.
Weak to moderate positive or negative correlations are observed between specific clusters.
For example, positive correlations are observed between C0 and C3 (0.41), whereas negative correlations are observed between C0 and C6 (-0.39).
This means that users belonging to these clusters have similar or different humor preferences.
These correlation results indicate differences in humor preferences among user clusters, further supporting the subjectivity of humor preferences.

\subsubsection{RQ2: Humor Preference Differences between LLMs and User Clusters\label{section:analysing_humor_preference_differences_between_llms_and_humans}}
Figure \ref{fig:all-users-vs-llm-btl-score-heatmap-top10-bottom10} summarizes the BTL scores of humor preference factors for the user clusters and three LLMs under the \texttt{no\_persona} setting.
To address RQ2, this visualization enables a direct comparison of humor preferences across models and user clusters.
Our key findings are as follows:

\paragraph{LLMs and User Clusters Exhibit Differences and Similarities in Humor Preference Factors}
Based on Figure \ref{fig:all-users-vs-llm-btl-score-heatmap-top10-bottom10}, we compare the BTL scores of humor preference factors between LLMs and user clusters.
Overall, both differences and similarities in BTL scores exist between LLMs and user clusters across multiple factors.
First, regarding the differences in preference factors between LLMs and user clusters, LLMs exhibited higher BTL scores than user clusters for overly long responses (\texttt{len-char-xlong}), responses with high vocabulary diversity (\texttt{vocab-unique-most}), and the use of slang (\texttt{slang}).
Second, regarding the similarities between LLMs and user clusters, both LLMs and users commonly favor factors such as \texttt{wordplay} and \texttt{dialogue}, whereas commonly disfavored factors include responses with a high proportion of nouns (\texttt{pos-noun-dominant}) and surreal/nonsense-style humor (\texttt{surreal\_nonsense}).
Furthermore, focusing on specific clusters, we observed that only certain clusters exhibit humor preferences similar to those of LLMs.
For instance, C5 shows high BTL scores for slang, as do LLMs; 
C0 also exhibits relatively high BTL scores for \texttt{len-token-xlong}, similar to LLMs.
These results suggest that LLMs share humor preferences with specific user clusters.

\paragraph{LLM Humor Preferences Similar to Specific User Clusters but not to Overall User Preferences}
To quantify preference similarity between user clusters and LLMs, we computed Pearson correlations between their BTL scores and humor preference factors.
Figure \ref{fig:user-cluster-btl-pearson} shows the Pearson's correlation coefficients.
Moderate positive correlations exist in humor preferences between LLMs and specific user clusters.
For example, LLMs and cluster C0 exhibit moderately positive correlations (GPT-5.1:0.57; Claude: 0.52).
Additionally, we observed weak negative correlations between the BTL scores of all users (estimated using vote data from all users without clustering) and those of each LLM (GPT-5.1: -0.22, Gemini: -0.36, Claude: -0.26).
These results not only support prior findings that LLMs possess humor preferences that differ from those of humans~\cite{sakabe2025assessingcapabilitiesllmshumora} but also provide new findings suggesting that LLMs may share humor preferences with specific user clusters.
We believe that these insights contribute to a deeper understanding of the relationship between LLM and human humor preferences.

\subsubsection{RQ3: Persona Prompting for Aligning LLM Preferences with User Clusters\label{section:analyzing_persona_prompting_efffect_on_LLM_preferences}}
\looseness=-1
To answer RQ3, we investigated whether persona prompting influences the alignment of LLMs' preference with specific clusters.
The key findings are as follows:

\paragraph{Persona Prompting Helped LLMs Align Humor Preferences with Specific User Clusters}
\begin{figure}[t]
  \centering
  \includegraphics[width=1\linewidth]{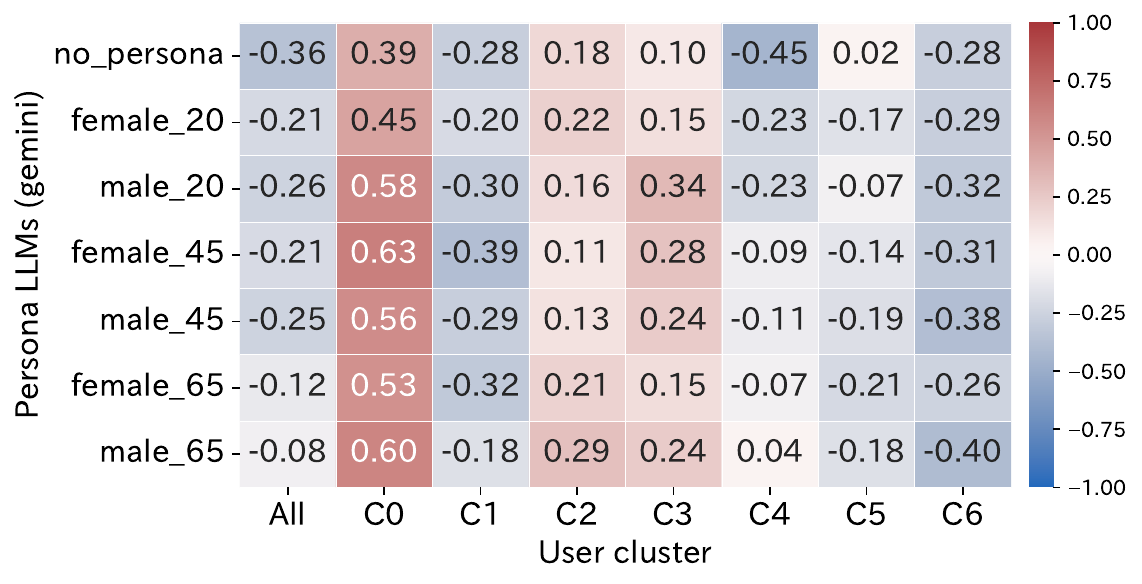}
  \caption{Pearson's correlation matrix between Gemini 3 Pro persona and user clusters, computed over the BTL scores of humor preference factors.}
  \label{fig:gemini-persona-vs-user-cluster-btl-pearson}
\end{figure}

\looseness=-1
We computed Pearson correlations between the BTL scores of humor preference factors for each persona-conditioned LLM and user cluster.
Figure~\ref{fig:gemini-persona-vs-user-cluster-btl-pearson} shows the resulting correlation matrix.
Persona prompting helped LLMs align their humor preferences with specific user clusters.
For instance, although weak positive correlations with C0 (0.39) are observed in the \texttt{no\_persona} setting, the \texttt{female\_45} persona exhibits a moderate positive correlation with C0 (0.63).
In another example, the \texttt{male\_20} persona exhibits a positive correlation with C3 (0.34), whereas the \texttt{no\_persona} setting exhibits a very weak positive correlation (0.10).
These results suggest that by assigning specific personas to LLMs, their humor preferences can be made to align with those of particular user clusters, paving the way for personalized optimization.\footnote{The true demographic attributes of each user cluster are unknown. Our claim is not that persona prompting replicates them but that it can align an LLM's humor preferences with specific user clusters.}

\section{Discussion\label{section:discussion}}
Building on our findings (\S\ref{section:humor-preference-factors-user-clusters}--\S\ref{section:analyzing_persona_prompting_efffect_on_LLM_preferences}), we discuss their broader implications and the interpretive scope of our analyses.

\paragraph{Preference Heterogeneity and Personalization}
Our preference comparison analysis between user clusters and the LLM suggests that, in subjective tasks with heterogeneous preferences, evaluating an LLM solely by its alignment with the overall user population is insufficient.
Because an LLM aligns with a specific cluster rather than the overall population, reporting alignment at the cluster (or individual) level clarifies whose preferences the LLM reflects.
This perspective also motivates personalized humor evaluation and generation that explicitly accounts for preference heterogeneity.
Developing such methods is an important direction for future work.

\paragraph{Effectiveness and Limitations of Persona Prompting}
We demonstrated that persona prompting can help LLMs align their humor preferences with those of specific user clusters.
This suggests that persona prompting may be effective in the context of humor preference, despite prior studies indicating its limited effect~\cite{zheng-etal-2024-helpful}.
However, persona prompting has certain limitations.
In our setting, persona prompting mainly changes alignment strength rather than switching the aligned cluster; see Appendix~\ref{appendix:persona_prompting_limitations}.
We leave individual-level alignment beyond persona prompting to future work.

\paragraph{Interpretable Analysis for Subjective Humor}
Humor is highly subjective, and this subjectivity is precisely why we designed our analysis to be interpretable, rather than relying only on aggregate scores or overall comparisons that are difficult to interpret: clustering summarizes groups of users with similar voting tendencies, and BTL modeling provides a factor-level view of which response features each group tends to prefer.
Our goal is not to define a single objective notion of humor quality, but to analyze differences in humor preferences across user groups and between humans and LLMs in an interpretable way.
A natural next step is to close the loop by validating these interpretable factors extrinsically, using the high-weight factors of each cluster as explicit guidance for cluster-targeted humor generation (see Limitations).

\paragraph{Exploratory Nature of Our Analyses}
Our clustering and factor-level BTL analyses are exploratory: the clusters are approximate groupings rather than cleanly separated user types (\S\ref{section:results_user_clustering}), and BTL scores should not be interpreted as independent causal effects of individual factors (\S\ref{sec:preference_modeling}).
Nevertheless, the random-partition baseline (Appendix~\ref{appendix:random_partition_baseline}) suggests that the observed cluster-level differences reflect non-random preference heterogeneity.
Building on this, future work could move from exploratory toward confirmatory analyses, for example by modeling factor interactions directly and testing specific cluster--factor associations on newly collected votes.

\section{Conclusion\label{section:conclusion}}
We analyzed heterogeneity in humor preferences in Japanese Oogiri by clustering users based on their voting histories and estimating cluster-level BTL scores over interpretable humor preference factors.
User clusters exhibited distinct preference factors (\S\ref{section:humor-preference-factors-user-clusters}); LLM preferences differed from those of the overall user population but resembled those of particular clusters (\S\ref{section:analysing_humor_preference_differences_between_llms_and_humans}); and persona prompting could direct LLM preferences toward specific clusters (\S\ref{section:analyzing_persona_prompting_efffect_on_LLM_preferences}).
Future work includes personalized humor generation and evaluation that explicitly model preference heterogeneity across users.

\section*{Limitations}
\paragraph{Limited Coverage of Humor Preference Factors}
The humor preference factors considered in this study are not exhaustive, and additional factors may influence humor preferences.
For instance, non-linguistic factors such as social and cultural background may affect humor preferences \cite{ruch1996cross,yue2016or,cao2023impact}, but we did not model them.
Future work should examine a broader set of humor preference factors.

\paragraph{Factor Dependencies and Interaction Effects}
Our factor-level BTL modeling prioritizes interpretability and does not model interaction effects or context-dependent humor.
Our correlation analysis (Appendix~\ref{appendix:factor_correlation}) showed that strong dependencies are not pervasive across the factor set, but some surface-form factors co-occur strongly; a high or low BTL score may thus partly reflect related response features, and individual factor weights should be interpreted with caution.

\paragraph{Extrinsic Validation via Factor-Guided Humor Generation}
Our study focuses on analyzing preference factors and does not validate them via downstream humor generation in the Oogiri setting.
Future work should test whether using the identified high-weight preference factors as guidance in the prompt can steer humor generation toward the preferences of specific user clusters, providing an extrinsic validation of our conclusions.

\paragraph{Focus on Text-based Oogiri Humor}
We focused on text-based Oogiri humor and did not consider multimodal humor content such as images or audio \cite{hossain-etal-2022-memosen,hessel-etal-2023-androids}.
Humor preferences for such content may differ from those for text.
Moreover, prior work suggested that even state-of-the-art LLMs have limited multimodal humor understanding, which raises concerns about the reliability of analyzing LLM humor preferences in multimodal settings \cite{zhou-etal-2025-bridging}.
Future work should extend our analysis to multimodal humor content.

\paragraph{Limited to Japanese Oogiri Humor}
Our analysis is based on Japanese Oogiri dataset \cite{murakami2025oogiricorpus}.
Because we focus on Japanese, our factors include language-specific characteristics such as character-type ratios.
Consequently, our findings may not directly transfer to other languages or cultural contexts.
However, we note that our user-cluster-based analysis method is language-agnostic and can be applied to other languages and cultures.
Broadening the data sources across languages and cultures is an important direction for generalization.

\paragraph{Limited Set of LLMs}
We analyzed three LLMs (GPT-5.1, Gemini-3-Pro, and Claude-Sonnet-4.5), selected based on reports that they exhibit stronger humor understanding than other open-source models \cite{murakami2025oogiricorpus}.
However, many other LLMs may have different humor preferences.
Future work should analyze a more diverse set of models, including language-specialized models and models with a wide range of parameter sizes.

\paragraph{Unknown True Attributes of Raters in Source Data}
Our clustering and humor preference factor analyses rely on user voting data collected from an Oogiri platform.
However, the platform does not provide verified information about raters' attributes or backgrounds.
Therefore, we cannot trace how each user cluster corresponds to raters' true personas or demographic attributes.
Collecting such information and clarifying these correspondences is an important direction for both personalized humor modeling and the interpretation of our analyses.
In future work, we plan to analyze how user information such as demographics and personality traits (e.g., the Big Five) relates to cluster membership and humor preferences, once such data become available.

\paragraph{Generalizability to Less Active Users}
To stabilize user vectors, we restrict the analysis to users with at least 100 votes (\S\ref{section:dataset_construction}). This focus on highly active users may limit the generalizability of our clusters and analyses to low-activity and cold-start users.
Developing methods that can robustly model users with sparse feedback remains an important direction for future work.

\paragraph{Sensitivity to Missing Votes in Clustering}
The vote data are inherently sparse because each user votes on only a subset of prompts.
As a result, user similarity estimation and clustering rely on partially overlapping observations, and the resulting clusters may be sensitive to how missing values are treated.
Moreover, the missingness can be informative (e.g., users may choose not to vote on prompts they dislike), potentially conflating preference similarity with participation patterns.
Future work should incorporate missingness-aware modeling and conduct robustness checks under alternative treatments of missing values, such as excluding missing values or imputing missing votes with a neutral value.

\paragraph{LLM vs Human Voting Setup}
On the Oogiri platform, humans may cast up to three votes per prompt, whereas our LLM task asks for a single funniest response per API call (we call the API three times and treat each as an independent trial).
This difference in the task setup may affect the comparability of LLM and human preference estimates; future work could align the setup (e.g., allowing the LLM to select up to three responses per prompt).

\paragraph{Prompt Dependency of LLM Preference Results}
LLM humor preference estimates may depend on the prompt template used for the funniest-response selection task (e.g., the no\_persona template underlying the main LLM-vs-cluster comparisons).
Our persona prompting experiments can be viewed as a sensitivity analysis over different prompts: we observed changes in the magnitude of correlations with user clusters but not a full reversal of preferences, suggesting that the reported LLM preferences are not entirely brittle to prompt wording.
Nevertheless, results under a single template should be interpreted with the possibility of prompt dependence in mind.

\paragraph{Reliability and Variance of Humor Strategy Labels}
We used an LLM to annotate humor strategy labels (\S\ref{section:humor_strategy_labels}) and validated them via a crowdsourced study (1,100 instances, five annotators per instance; 81.4\% judged appropriate overall).
Nevertheless, quality varies across labels (e.g., \texttt{incongruity} has lower accuracy), and the crowdsourced evaluation does not guarantee that every label is equally reliable across the full dataset.
Results that depend on these labels should be interpreted with this variance in mind.

\section*{Ethical Considerations}
\paragraph{Data Source}
The dataset used in our study is a Japanese Oogiri dataset collected by \citet{murakami2025oogiricorpus}.
We used the dataset in accordance with its intended use and license (CC BY-NC-SA 4.0).
The dataset was collected from a public website in compliance with its terms of use, and the site permits automated crawling as specified in \texttt{robots.txt}.
To support reproducibility, we release the code for data collection, processing, and analysis under the MIT license.\footnote{\url{https://github.com/CyberAgentAILab/oogiri-dataset-builder}}

\paragraph{Privacy}
Although the data are publicly available, they may still contain personal information such as usernames or other identifiers.
In our analysis, we took care to protect user privacy.
We removed or anonymized identifiers and avoided including content that could enable re-identification.

\paragraph{Potential Risks}
Although our study is intended for research purposes, analyses of user preferences may be misused for profiling or for drawing unwarranted inferences about individuals or groups. There is also a residual risk of re-identification when working with public web data. To mitigate these risks, we (i) anonymized identifiers and do not release personal identifiers, (ii) focus on aggregate, cluster-level analyses rather than individual-level claims, and (iii) explicitly limit the scope of our conclusions as discussed in the Limitations section.

\paragraph{Bias and Generalizability}
Because our data are Japanese and culture-dependent, our findings may not generalize to other languages or cultural contexts as we discussed in the Limitations section.
The observed preferences may also reflect societal biases present in the source platform.

\paragraph{Use of LLMs for Writing Assistance}
We used an LLM as a writing assistant for English proofreading and editing.
All technical content, analyses, and conclusions were produced and verified by the authors.

\bibliography{references}

\clearpage

\appendix

\section{Clustering Robustness Analyses\label{appendix:clustering_robustness}}
To assess whether our user clusters are reproducible rather than artifacts of a single setup, we report three robustness analyses: seed sensitivity, comparison with other clustering methods, and sensitivity to the number of clusters $K$.
We treated the baseline partition as the reference and measured agreement using the Adjusted Rand Index (ARI) and Normalized Mutual Information (NMI).

\paragraph{Implementation Details}
We used scikit-learn's \texttt{KMeans} with k-means++ initialization and \texttt{n\_init}=10, and we used \texttt{random\_state}=42 to define the baseline partition.
The number of clusters was set to $K=7$ based on the elbow method~\cite{thorndike1953belongs_elbow_method} and silhouette scores~\cite{Rousseeuw1987-silhouette}.

\paragraph{Selection of $K$}
Table~\ref{tab:appendix-clustering-k-selection} reports inertia and silhouette scores for $K=2$ to $19$.
$K$=7 achieved the highest silhouette score in this range (0.0258), although $K$=10 was very close (0.0257), and inertia did not show a single decisive elbow.
We selected $K$=7 as a practical trade-off: larger $K$ values make cluster-level BTL estimation more sparse, whereas smaller $K$ values may merge users with different preferences into the same cluster.
We thus present $K$=7 as a reasonable analytical choice for this exploratory study rather than as a uniquely optimal number of clusters.

\begin{table}[t]
\centering
\small
\begin{tabular}{c r r}
\toprule
\textbf{$K$} & \textbf{Inertia} & \textbf{Silhouette} \\
\midrule
2 & 261.62 & 0.0190 \\
3 & 257.23 & 0.0207 \\
4 & 254.03 & 0.0208 \\
5 & 250.51 & 0.0233 \\
6 & 247.03 & 0.0239 \\
\textbf{7} & \textbf{244.60} & \textbf{0.0258} \\
8 & 242.18 & 0.0245 \\
9 & 240.21 & 0.0247 \\
10 & 237.62 & 0.0257 \\
11 & 236.29 & 0.0252 \\
12 & 234.25 & 0.0227 \\
13 & 232.92 & 0.0238 \\
14 & 231.88 & 0.0226 \\
15 & 230.35 & 0.0199 \\
16 & 228.39 & 0.0205 \\
17 & 226.68 & 0.0202 \\
18 & 224.84 & 0.0195 \\
19 & 224.05 & 0.0202 \\
\bottomrule
\end{tabular}
\caption{Inertia and silhouette scores for different numbers of clusters $K$. $K$=7 (in bold) achieved the highest silhouette score in this range, although $K$=10 was very close (0.0257).}
\label{tab:appendix-clustering-k-selection}
\end{table}

\paragraph{Seed Sensitivity}
We ran K-means with 100 different random seeds (varying \texttt{random\_state}) and computed ARI and NMI of each resulting partition against the baseline.
Table~\ref{tab:appendix-clustering-seed-stability} reports the mean, standard deviation, and median.
The median ARI (0.420) and NMI (0.555) indicate moderate agreement across initializations.

\begin{table}[t]
\centering
\small
\begin{tabular}{l r r r}
\toprule
\textbf{Metric} & \textbf{Mean} & \textbf{Std} & \textbf{Median} \\
\midrule
ARI  & 0.420 & 0.076 & 0.420 \\
NMI  & 0.558 & 0.055 & 0.555 \\
\bottomrule
\end{tabular}
\caption{Stability of user clustering over 100 random seeds. Agreement (ARI, NMI) with the baseline partition (random\_state=42) is reported.}
\label{tab:appendix-clustering-seed-stability}
\end{table}

\paragraph{Comparison with Other Methods}
We compared the baseline K-means partition with Agglomerative and Spectral clustering (same user embeddings and $K$=7, cosine similarity).
Table~\ref{tab:appendix-clustering-method-comparison} shows that both methods achieve moderate agreement (ARI $\approx$ 0.42-0.43, NMI $\approx$ 0.52-0.58) with the baseline, suggesting that the main partition structure is not specific to K-means.

\begin{table}[t]
\centering
\small
\begin{tabular}{l r r}
\toprule
\textbf{Method} & \textbf{ARI} & \textbf{NMI} \\
\midrule
Agglomerative & 0.430 & 0.523 \\
Spectral       & 0.417 & 0.581 \\
\bottomrule
\end{tabular}
\caption{Agreement with the baseline K-means partition ($K$=7). User embeddings and $K$=7 were kept the same; proximity was cosine similarity.}
\label{tab:appendix-clustering-method-comparison}
\end{table}

\paragraph{Sensitivity to $K$}
We varied $K$ to 6 and 8 and compared the resulting partitions to the baseline ($K$=7) using ARI and NMI (Table~\ref{tab:appendix-clustering-k-sensitivity}).
Although some users shift clusters when $K$ changes, NMI remains around 0.53--0.54 for $K \neq 7$, indicating that the overall structure is not extremely sensitive to small changes in $K$.
Note that these sensitivity analyses are not intended as the sole justification for selecting $K$=7 (see the Selection of $K$ paragraph above); rather, they show that the overall partition structure is robust to nearby choices of $K$.

\begin{table}[t]
\centering
\small
\begin{tabular}{c r r}
\toprule
\textbf{$K$} & \textbf{ARI} & \textbf{NMI} \\
\midrule
6 & 0.347 & 0.534 \\
7 & 1.000 & 1.000 \\
8 & 0.378 & 0.543 \\
\bottomrule
\end{tabular}
\caption{Sensitivity to the number of clusters. ARI and NMI are computed against the baseline partition with $K$=7.}
\label{tab:appendix-clustering-k-sensitivity}
\end{table}

\section{Random-Partition Baseline for Cluster-Level BTL Scores\label{appendix:random_partition_baseline}}
To test whether the BTL-score differences observed across user clusters (\S\ref{section:humor-preference-factors-user-clusters}) are genuinely driven by divergent humor preferences rather than by vote sparsity or the sensitivity of the estimation, we compared the K-means clusters against a random-partition baseline.

\paragraph{Setup}
For the K-means clusters, we computed, for each factor, the variance of its cluster-level BTL scores across the seven clusters, and averaged these variances over all factors; we refer to this metric as the mean factor-wise BTL-score variance.
For the random baseline, we shuffled all users and partitioned them into seven groups with the same sizes as the K-means clusters, so that each user appeared in exactly one random group.
We repeated this procedure over 30 random partitions, re-estimated factor-level BTL scores for each random group using the same pipeline, and computed the same metric.

\paragraph{Results}
As shown in Table~\ref{tab:appendix-random-partition-baseline}, the mean factor-wise BTL-score variance was higher for the K-means clusters than for the random baseline (0.0145 vs. 0.0083 $\pm$ 0.0010): the same factors received more divergent BTL scores across the K-means clusters than across random same-size groups.
None of the 30 random partitions exceeded the K-means value; a one-sided empirical test with the standard +1 correction yields $p = 0.032$.
This suggests that the observed cluster-level BTL-score differences are unlikely to arise simply from splitting users into smaller groups of the same sizes, and we therefore interpret the clusters as capturing non-random preference heterogeneity.
At the same time, these clusters remain exploratory groupings for analysis; we do not claim that humor preferences are cleanly divided into seven discrete user types.

\begin{table}[t]
\centering
\small
\begin{tabular}{l r}
\toprule
\textbf{Partition} & \textbf{Mean variance} \\
\midrule
K-means clusters ($K$=7) & 0.0145 \\
Random partitions (30 runs) & 0.0083 $\pm$ 0.0010 \\
\bottomrule
\end{tabular}
\caption{Random-partition baseline for the mean factor-wise variance of cluster-level BTL scores. The random baseline reports the mean and standard deviation over 30 same-size random partitions. None of the 30 random partitions exceeded the K-means value (one-sided empirical test with the +1 correction, $p = 0.032$).}
\label{tab:appendix-random-partition-baseline}
\end{table}

\section{Humor Preference Factors\label{appendix:definition_of_humor_preference_factors}}
This section provides the detailed definitions of the humor preference factors used in this study.
The humor preference factors are broadly categorized into linguistic features (\S\ref{appendix:definition_of_linguistic_features}) and humor strategy labels (\S\ref{appendix:humor_strategy_labeling}).

\subsection{Linguistic Features\label{appendix:definition_of_linguistic_features}}

\begin{table*}
\centering
{\small
\begin{tabularx}{\textwidth}{lX}
\toprule
\multicolumn{1}{c}{\textbf{Feature Name}} & \multicolumn{1}{c}{\textbf{Definition}} \\
\midrule

\multicolumn{2}{l}{\textbf{Basic Features (11 features)}} \\
len-char-\{short, medium, long, xlong\} & Number of characters in the response (quartile-based) \\
hiragana-\{minimal, low, medium, high\} & Ratio of hiragana characters (quartile-based) \\
katakana-\{minimal, high\} & Ratio of katakana characters (median-based, 2 levels) \\
kanji-\{minimal, low, medium, high\} & Ratio of kanji characters (quartile-based) \\
alphabet-\{minimal, high\} & Ratio of alphabet characters (median-based, 2 levels) \\
digit-\{minimal, high\} & Ratio of digit characters (median-based, 2 levels) \\
punct-\{minimal, high\} & Ratio of punctuation marks (median-based, 2 levels) \\
space-\{minimal, high\} & Ratio of space characters (median-based, 2 levels) \\
symbol-\{minimal, high\} & Ratio of symbol characters (median-based, 2 levels) \\
punct-count-\{few, most\} & Number of punctuation marks (median-based, 2 levels) \\
sentences-\{one, many\} & Number of sentences (median-based, 2 levels) \\
\midrule

\multicolumn{2}{l}{\textbf{Morphological Analysis Features (10 features)}} \\
pos-noun-\{low, medium, high, dominant\} & Ratio of nouns (quartile-based) \\
pos-verb-\{minimal, high\} & Ratio of verbs (median-based, 2 levels) \\
pos-adj-\{minimal, high\} & Ratio of adjectives (median-based, 2 levels) \\
pos-adverb-\{minimal, high\} & Ratio of adverbs (median-based, 2 levels) \\
pos-particle-\{minimal, low, medium, high\} & Ratio of particles (quartile-based) \\
pos-auxiliary-\{minimal, high\} & Ratio of auxiliary verbs (median-based, 2 levels) \\
len-token-\{short, medium, long, xlong\} & Number of tokens (quartile-based) \\
vocab-unique-\{few, some, many, most\} & Number of unique tokens (quartile-based) \\
vocab-diversity-\{repetitive, very-diverse\} & Lexical diversity: unique tokens / total tokens (median-based, 2 levels) \\
proper-noun & Contains proper nouns \\
\midrule

\multicolumn{2}{l}{\textbf{Special Symbol Features (5 features)}} \\
dialogue & Contains dialogue markers (\ja{「」}) \\
parentheses & Contains parentheses \\
tilde & Contains tilde (\textasciitilde) \\
number & Contains numbers \\
slang & Contains internet slang (www) \\
\midrule

\multicolumn{2}{l}{\textbf{Sentence-Ending Pattern Features (9 features)}} \\
ending-period & Ends with a period \\
ending-question & Ends with a question mark \\
ending-exclamation & Ends with an exclamation mark \\
ending-ellipsis & Ends with an ellipsis \\
ending-noun & Ends with a noun \\
ending-verb & Ends with a verb \\
ending-adjective & Ends with an adjective \\
ending-particle & Ends with a particle \\
ending-auxiliary & Ends with an auxiliary verb \\
\midrule

\multicolumn{2}{l}{\textbf{Writing Style Features (4 features)}} \\
polite-style & Polite style (desu/masu form) \\
casual-style & Casual style (da form) \\
exaggeration-rule & Contains exaggeration expressions \\
negation & Contains negation expressions \\
\midrule

\multicolumn{2}{l}{\textbf{Prompt-Response Relational Features (6 features)}} \\
prompt-overlap-\{minimal, low, medium, high\} & Character overlap ratio between prompt and response (quartile-based) \\
prompt-kanji-share-\{minimal, high\} & Shared kanji ratio between prompt and response (median-based, 2 levels) \\
prompt-noun & Contains nouns from the prompt \\
prompt-verb & Contains verbs from the prompt \\
prompt-proper-noun & Contains proper nouns from the prompt \\
length-ratio-\{short, balanced, long, very-long\} & Ratio of response length to prompt length (quartile-based) \\
\bottomrule
\end{tabularx}}
\caption{Definition of linguistic features.}
\label{tab:definition_of_linguistic_features}
\end{table*}

\subsubsection{Definition}
Table \ref{tab:definition_of_linguistic_features} summarizes the detinitions of the linguistics features.
The linguistic features consist of six subgroups: basic features, morphological analysis features, special symbol features, sentence-ending pattern features,  writing-style features, and prompt-response relational features, totaling 45 features.
The first five subgroups are based on the linguistic characteristics of the responses, while the last subgroup is based on the relationship between the prompt and the response.
We describe each subgroup in detail below.

\paragraph{Basic Features (11 features)}
This feature subgroup includes basic linguistic features such as the number of characters in the response and the ratios of different character types.
Specifically, it comprises the following features: character count; the hiragana, katakana, kanji, alphabet, digit, punctuation, space, and symbol ratios; punctuation and sentence counts.

\paragraph{Morphological Features (10 features)}
This feature subgroup includes features derived from the morphological analysis of the responses, such as POS ratios and token counts.
It consists of the following features: noun, verb, adjective, adverb, particle, and auxiliary verb ratios; token and unique token counts; lexical diversity (unique words/total words); and the presence of proper nouns.

\paragraph{Special Symbol Features (5 features)}
This subgroup includes features based on the presence of specific symbols or patterns in a response.
Specifically, it consists of indicators of whether the response contains dialogue quotes (\ja{「」}), parentheses, tilde ($\sim$), numbers, internet slang phrases such as ``www'' or ``\ja{草}''.
The internet slang phrases ``www'' and ``\ja{草}'' are commonly used in Japanese online communities, similar to ``lol'' in English.

\paragraph{Sentence-Ending Pattern Features (9 features)}
This feature subgroup includes features based on POS, punctuation marks, or symbols that appear at the end of a response.
Specifically, it consists of indicators of whether a response ends with a period (.), question mark (?), exclamation mark (!) or ellipsis ($\ldots$) and whether it ends with a noun, verb, adjective, particle, or auxiliary verb.

\paragraph{Writing Style Features (4 features)}
This subgroup includes features based on the presence of specific writing styles unique to the Japanese language, such as polite, casual, exaggerated, and negated forms.

\paragraph{Prompt-Response Relational Features (6 features)}
This feature subgroup consists of features based on the relationship between the prompt and response.
Specifically, it includes the following features: character overlap ratio between the prompt and response, shared kanji ratio between the prompt and response, response-to-prompt length ratio, and indicators of whether the response contains nouns, verbs, or proper nouns from the prompt.

\subsubsection{Binning of Continuous Features}\label{appendix:binning_countinous_features}
To estimate the relative strength of each factor in the BTL model, all features must be treated as categorical factors.
In this study, following the DecipherPref framework \cite{hu-etal-2023-decipherpref}, we apply binning to convert continuous-valued features into categorical factors.
Specifically, we use quartile-based binning (4 levels) for continuous-valued features.
Features that cannot be reliably categorized based on quartiles (e.g., due to many duplicated values and insufficient value diversity) are split into two levels using the median.

For example, the feature \texttt{len-char} (the number of characters in a response) is binned into four quartile-based categories (\texttt{len-char-short}, \texttt{len-char-medium}, \texttt{len-char-long}, \texttt{len-char-xlong}), whereas \texttt{pos-verb} (the ratio of verbs in a response) is binned into two median-based categories (\texttt{pos-verb-minimal}, \texttt{pos-verb-high}).
Details of the binning scheme for each feature are provided in Table \ref{tab:definition_of_linguistic_features}.

\subsubsection{Implementation}
For extracting linguistic features, we used SudachiPy \cite{takaoka-etal-2018-sudachi} for morphological analysis, and Python's standard library \texttt{unicodedata} and regular expressions for character type identification.
SudachiPy provides morpheme segmentation and part-of-speech tags, which we use to compute POS-based features (e.g., POS ratios and ending POS indicators) for each response.
We used \texttt{unicodedata} and regular expressions to categorize character types and detect punctuation/symbol markers; for relational features, we applied the same procedure to both the prompt and response texts.
For all tools, we used the default settings.

\subsection{Humor Strategy Labels\label{appendix:humor_strategy_labeling}}
\begin{figure}[t]
  \centering
  \begin{lstlisting}[
    basicstyle=\ttfamily\scriptsize\linespread{0.75}\selectfont,
    breaklines=true,
    breakatwhitespace=true,
    aboveskip=0pt,
    belowskip=0pt,
    backgroundcolor=\color{gray!10},
    frame=single,
    framerule=0.4pt,
    framesep=4pt,
    rulecolor=\color{gray!40}
  ]
You are an annotator for Ogiri responses. Read the given prompt and response, and assign predefined response strategy labels accurately and consistently. Think carefully so you can explain all decision rationale, and select a confidence level that represents the reliability of your judgment.

## Task Overview
Read the given Ogiri prompt and response pairs, and based on the strategy label definitions below, annotate each response with the most appropriate strategy label. Multiple label assignments are permitted. Labels must be assigned based on the label definitions.

## Strategy Label Definitions
- Label: {{label_name}}
  Definition: {{definition}}
  Guidelines: {{guidelines}}
  Ambiguity Notes: {{ambiguity_notes}}
  Examples: {{examples}}

... (The remaining label definitions are omitted for brevity.)

## Confidence Levels
- Available confidence: high, medium, low

## Output Format
{
  "items": [
    {
      "prompt_id": "string",
      "response_id": "string",
      "selected_labels": [
        {
          "reason": "string",
          "label": "string",
          "confidence": "string"
        }
      ],
    }
  ],
}

## Annotation Target
prompt_id: {{prompt_id}}
response_id: {{response_id}}
Prompt: {{prompt}}
Response: {{response}}
...
  \end{lstlisting}
  \caption{Prompt template for humor strategy labeling.}
  \label{fig:prompt-for-humor-strategy-labeling}
\end{figure}

\begin{table*}[htbp]
\centering

\small
\setlength{\tabcolsep}{4pt}
\begin{tabularx}{\textwidth}{@{}>{\raggedright\arraybackslash}p{3cm} >{\raggedright\arraybackslash}X >{\raggedright\arraybackslash}X >{\raggedright\arraybackslash}X@{}}
\toprule
\textbf{Label Name} & \textbf{Definition} & \textbf{Guidelines} & \textbf{Ambiguity Notes} \\
\midrule

\texttt{wordplay} &
Manipulates surface linguistic features (sound, characters, syntax) to create humor through puns, double meanings, or rhythm. &
Check for phonetic substitutions, puns, or rhythmic structure. Use this label when surface-level manipulation is primary. &
If the humor comes solely from meaning inversion, consider \texttt{incongruity}. Focus on ``surface-level linguistic manipulation.'' \\

\midrule

\texttt{shared\_experience} &
Draws on shared everyday experiences, eliciting laughter through empathy. &
Confirm empathy is primary; \texttt{exaggeration} is secondary. &
If the humor relies primarily on impact, use \texttt{exaggeration}. \\

\midrule

\texttt{exaggeration} &
Exaggerates or downplays quantity, emotion, or scale to an extreme. &
Determine whether exaggeration is the primary goal. &
If empathy is core, use \texttt{shared\_experience}. \\

\midrule

\texttt{black\_joke\_satire} &
Engages with social norms or taboos, creating humor through irony and taboo language. &
Record the source and target categories (i.e., what is being referenced vs. what is being targeted). Distinguish it from exaggeration and incongruity. &
If the humor primarily stems from the prompt's absurdity, consider \texttt{surreal\_nonsense}. If it relies only on expectation reversal, consider \texttt{incongruity}. \\

\midrule

\texttt{surreal\_nonsense} &
Severs contextual connections, making absurdity itself the source of humor. &
Confirm that the logical leap is intentional. &
Context destruction is required; eccentricity alone is insufficient. \\

\midrule

\texttt{incongruity} &
Uses a reversal of an expected development/premise as the punchline. &
Confirm that the response sets up an expectation and then reverses it. Check whether the reason for the reversal is explicit. &
If the target is the oogiri framework or the prompt itself, use \texttt{meta}. If ethical criticism is primary, consider \texttt{black\_joke\_satire}. \\

\midrule

\texttt{meta} &
Refers to contradictions in the oogiri framework, its rules, or the prompt itself, creating humor from an external perspective. &
Check whether framework elements (recording, host, format) are used. Use when pointing out flaws in the prompt. &
If the humor is purely meaning inversion or expectation violation, consider \texttt{incongruity}. If it includes an overview of the framework, prioritize \texttt{meta}. \\

\midrule

\texttt{self\_reference} &
Uses the responder's own shortcomings as material, creating humor from a first-person perspective. &
Check for first-person pronouns and whether the responder's own characteristics/failures are central to the punchline. &
If the response primarily criticizes the framework or the prompt, prioritize \texttt{meta}. \\

\midrule

\texttt{personification} &
Gives human-like emotions or a voice to inanimate objects, creating humor through characterization. &
Check whether the target is clearly personified. Judge whether the character's voice is the primary driver of the humor. &
If borrowing entire settings/stories, use \texttt{parody} as primary, \texttt{personification} as secondary. \\

\midrule

\texttt{parody} &
Borrows or transforms settings/stories from external content, creating humor through the gap between the response and the source material. &
Explicitly identify the source material. Confirm structural borrowing (not just proper nouns). &
Proper nouns alone do not constitute parody. Use structural, setting, or story borrowing as the criterion. \\

\midrule

\texttt{mini\_story} &
Depicts a short story/scene, creating humor in the conclusion. &
Check whether a specific situation/scene is described and ends with a punchline. Confirm that the narrative structure is primary. &
If story elements are primary, use \texttt{mini\_story} as the main label. If the response is a one-liner or primarily \texttt{wordplay}, prioritize other labels. \\

\bottomrule
\end{tabularx}
\caption{Humor strategy labels: definitions, guidelines, and ambiguity notes\label{tab:humor_strategy_labels}}
\end{table*}

\begin{table}[t]
\centering
\small
\setlength{\tabcolsep}{4pt}
\begin{tabular}{l r r}
\toprule
\textbf{Humor strategy label} & \textbf{Accuracy} & \textbf{Agreement rate} \\
\midrule
\texttt{shared\_experience}    & 0.880 & 0.774 \\
\texttt{black\_joke\_satire}   & 0.820 & 0.784 \\
\texttt{exaggeration}          & 0.700 & 0.704 \\
\texttt{incongruity}           & 0.590 & 0.716 \\
\texttt{meta}                  & 0.830 & 0.734 \\
\texttt{mini\_story}           & 0.910 & 0.762 \\
\texttt{parody}                & 0.830 & 0.778 \\
\texttt{personification}       & 0.840 & 0.778 \\
\texttt{self\_reference}       & 0.790 & 0.794 \\
\texttt{surreal\_nonsense}     & 0.890 & 0.804 \\
\texttt{wordplay}              & 0.870 & 0.752 \\
\midrule
\textbf{Overall}               & 0.814  & 0.762  \\
\bottomrule
\end{tabular}
\caption{Crowdsourced quality evaluation of humor strategy labels (1,100 instances, 5 annotators per instance, majority vote). Accuracy: proportion judged valid by majority. Agreement rate: average proportion of annotators supporting the majority decision.}
\label{tab:appendix-quality-eval-humor-strategy-label-crowd}
\end{table}

\begin{table*}[t]
\centering
\small
\setlength{\tabcolsep}{4pt}
\begin{tabularx}{\textwidth}{@{}>{\raggedright\arraybackslash}X >{\raggedright\arraybackslash}X >{\raggedright\arraybackslash}p{3.1cm}@{}}
\toprule
\textbf{Prompt} & \textbf{Response} & \textbf{Label} \\
\midrule
Build suspense in one line. & You know why you were called in, right? & \texttt{shared\_experience} \\
Tell me the most pointless trivia in the world. & I am in a bad mood when I wake up. & \texttt{self\_reference} \\
What is a wedding you would hate like? & It is held every day. & \texttt{exaggeration} \\
Slimy horse racing: what is it like? & Every horse is newborn. & \texttt{surreal\_nonsense} \\
We are recruiting heroes. What are the eligibility requirements? & Someone whose hometown was destroyed. & \texttt{incongruity} \\
\bottomrule
\end{tabularx}
\caption{Annotation examples of humor-strategy labels. For visibility, Japanese prompts and responses are translated into English.}
\label{tab:appendix-examples-humor-strategy-labels}
\end{table*}

This section provides the detailed definitions of the humor strategy labels, the implementation of feature extraction, and the quality evaluation of the annotated labels.
\subsubsection{Definition and Prompt Template}
Figure \ref{fig:prompt-for-humor-strategy-labeling} shows the prompt template for humor strategy labeling, and Table \ref{tab:humor_strategy_labels} presents the definitions of each label, annotation guidelines, and instructions for handling ambiguous cases.
The {\ttfamily \{\{label\_name\}\}}, {\ttfamily \{\{definition\}\}}, {\ttfamily \{\{guidelines\}\}}, and {\ttfamily \{\{ambiguity\_notes\}\}} parts of the prompt template are replaced with the label name, definition, annotation guidelines, and instructions for handling ambiguous cases for each label shown in Table \ref{tab:humor_strategy_labels}, respectively.
The {\ttfamily \{\{examples\}\}} part is replaced with specific examples of annotations corresponding to each label.
Additionally, the {\ttfamily \{\{prompt\_id\}\}}, {\ttfamily \{\{response\_id\}\}}, {\ttfamily \{\{prompt\}\}}, and {\ttfamily \{\{response\}\}} parts of the prompt template are replaced with the IDs and texts of the prompt and response, respectively.

\subsubsection{Implementation}
For humor strategy labeling, we used GPT-5.1, a state-of-the-art LLM, via an API.
For each API call, we set \texttt{temperature}=1 and used default values for all other parameters.
To mitigate variability in API outputs, we followed a self-consistency protocol \cite{wang-etal-2023-self-consistency}: for each labeling prompt, we collected three responses and selected the final label by majority voting.
In addition, because the annotation set was large (14,389 instances), we adopted a batch prompting strategy \cite{cheng-etal-2023-batch} to reduce API costs, processing 20 samples per API call.

\subsubsection{Quality Evaluation of Annotation Labels}
To evaluate the annotation quality of humor-strategy labels, we conducted a crowdsourced evaluation on 1,100 instances (100 per label) via Yahoo! Crowdsourcing, a major Japanese crowdsourcing platform.
Each instance was judged by five independent Japanese-speaking annotators; the final label was determined by majority vote.
Annotators were shown the prompt, response, humor strategy label, and label definitions, and answered a binary question: \textit{Is the humor-strategy label appropriate for this prompt-response pair?}
We administered a comprehension test on the label definitions; only workers who passed could participate.
Following the platform's payment guidelines, we set the reward to 10 yen per 25 tasks.
Workers were notified that their annotations would be used for research.

Overall, 81.4\% of the 1,100 instances were judged appropriate (average agreement rate 76.2\%).
Table~\ref{tab:appendix-quality-eval-humor-strategy-label-crowd} reports accuracy and agreement rate per label; Table~\ref{tab:appendix-examples-humor-strategy-labels} provides annotation examples.
Quality varies across labels (e.g., \texttt{incongruity} has relatively lower accuracy), which we attribute to definition ambiguity; results should be interpreted with this variance in mind (see Limitations).
In future work, improving label definitions and validation with additional annotators will be important.

\section{Correlation Analysis of Humor Preference Factors\label{appendix:factor_correlation}}
Humor is often shaped by combinations of response features, context, and overall expression; therefore, factor-level BTL scores should not be interpreted as independent causal effects of isolated features.
To examine dependencies among the factors, we conducted a co-occurrence/correlation analysis using the same 97 factor values as in the BTL estimation.
Each response was represented as a multi-hot vector over these factor values, and we computed pairwise phi correlations between factor indicators.
We focused on cross-category factor pairs, i.e., pairs of factor values that belong to different feature categories, because values within the same category are mutually exclusive by construction (e.g., a response cannot be both \texttt{len-token-short} and \texttt{len-token-long}).

As summarized in Table~\ref{tab:appendix-factor-correlation-summary}, most cross-category factor pairs showed weak correlations: across 4,591 pairs, the median absolute phi correlation was 0.033, the mean was 0.068, and the 90th percentile was 0.160, suggesting that strong dependencies are not pervasive across the factor set.
However, a small number of pairs showed strong correlations, mostly among mechanically related surface-form features such as token length, character length, vocabulary count, punctuation, and number-related indicators.
For example, \texttt{len-token-short} and \texttt{vocab-unique-few} were strongly correlated ($\phi = 0.973$), as were \texttt{len-token-xlong} and \texttt{vocab-unique-most} ($\phi = 0.869$): shorter responses naturally tend to contain fewer unique tokens, whereas very long responses tend to contain more unique tokens.
Accordingly, factor-level BTL scores indicate which response features tend to be preferred under our modeling setup, but they should not be interpreted as independent causal effects of individual factors: because some factors co-occur, a high or low BTL score may partly reflect related response features.

\begin{table}[t]
\centering
\small
\begin{tabular}{l r}
\toprule
\textbf{Statistic} & \textbf{Value} \\
\midrule
Factor values used in BTL estimation & 97 \\
Cross-category factor pairs analyzed & 4,591 \\
Median $|\phi|$ & 0.033 \\
Mean $|\phi|$ & 0.068 \\
90th percentile of $|\phi|$ & 0.160 \\
95th percentile of $|\phi|$ & 0.264 \\
\bottomrule
\end{tabular}
\caption{Summary of pairwise phi correlations between binary factor indicators over responses. The summary statistics are computed over cross-category factor pairs.}
\label{tab:appendix-factor-correlation-summary}
\end{table}

\section{Persona Prompting\label{appendix:overview_of_persona_prompting}}
This section describes the persona prompting template and detailed implementation.

\paragraph{Prompt Template}
\begin{figure}[t]
\centering
\begin{lstlisting}[
  basicstyle=\ttfamily\scriptsize\linespread{0.75}\selectfont,
  breaklines=true,
  breakatwhitespace=true,
  aboveskip=0pt,
  belowskip=0pt,
  backgroundcolor=\color{gray!10},
  frame=single,
  framerule=0.4pt,
  framesep=4pt,
  rulecolor=\color{gray!40}
]
{{ persona_description }}

Your role is to look at multiple responses to a given Ogiri prompt, and select the response you truly find funny based on your own unique sensibility. Your task is to choose the response you find funny.

## Evaluation Rules
1. Select at most 1 response that you find funny
2. If there are no responses you find funny, you must not select any. (0 selections is acceptable)
3. For the response you select, provide a brief explanation of why you found it funny

## Prompt
{{ prompt }}

## Response_candidates
{{ response_candidates }}

## Output Format
Please output in the following JSON format.
Do not include any text other than JSON.

{
  "selected_responses": [
    {
      "reasoning": "Reason for selecting that response (approximately 50 characters)"
      "response_id": "Response ID",
    }
  ]
}

Notes:
- Include only the selected response(s) in selected_responses (maximum 1)
- If there are no responses you find funny, set selected_responses to an empty array `[]`.
- Use the exact IDs listed in the response candidates above for response_id
\end{lstlisting}
\caption{Prompt template for persona prompting.}
\label{fig:prompt-for-persona-prompting}
\end{figure}

\begin{table*}[t]
  \centering
  \small
  \begin{tabularx}{\textwidth}{@{}>{\centering\arraybackslash}p{0.18\textwidth}X@{}}

    \toprule
    \multicolumn{1}{c}{\bf{Persona}} & \multicolumn{1}{c}{\bf{Description}} \\
    \midrule
    \multicolumn{1}{c}{\texttt{no\_persona}} & Please evaluate the Ogiri responses. \\
    \multirow{3}{*}{\texttt{female\_20}} &
    You are a 20-year-old female born in 2005.
    You are a university student who enjoys comedy variety shows.
    You are sensitive to cute and emotionally resonant things, and you like relatable content.
    Please evaluate the Ogiri responses with a young woman's sensibility. \\
    \multirow{3}{*}{\texttt{male\_20}} &
    You are a 20-year-old male born in 2005.
    You are a university student who frequently uses SNS and watches YouTube.
    You are well-versed in trending topics, memes, and internet slang.
    Please evaluate the Ogiri responses with a youthful sensibility. \\
    \multirow{4}{*}{\texttt{female\_45}} &
    You are a 45-year-old female born in 1980.
    You work as a company employee and have a family (husband and two children).
    You are knowledgeable about Showa and Heisei era comedy and current affairs.
    Please evaluate based on the common sense and experience you have cultivated as a working professional. \\
    \multirow{4}{*}{\texttt{male\_45}} &
    You are a 45-year-old male born in 1980.
    You work as a company employee and have a family (wife and two children).
    You are knowledgeable about Showa and Heisei era comedy and current affairs.
    Please evaluate based on the common sense and experience you have cultivated as a working professional. \\
    \multirow{3}{*}{\texttt{female\_65}} &
    You are a 65-year-old female born in 1959.
    After retirement, you enjoy pursuing hobbies.
    You enjoy traditional comedy such as rakugo and manzai.
    Please evaluate the Ogiri responses from a perspective enriched by life experience. \\
    \multirow{3}{*}{\texttt{male\_65}} &
    You are a 65-year-old male born in 1959.
    After retirement, you enjoy pursuing hobbies.
    You enjoy traditional comedy such as rakugo and manzai.
    Please evaluate the Ogiri responses from a perspective enriched by life experience. \\
    \bottomrule
  \end{tabularx}
  \caption{Persona descriptions used for persona prompting.}
  \label{tab:persona-descriptions}
\end{table*}

Figure \ref{fig:prompt-for-persona-prompting} shows the prompt template for persona prompting.
Table \ref{tab:persona-descriptions} lists the descriptions of each persona used in the prompt template.
For example, the {\ttfamily \{\{persona\_description\}\}} part of the prompt template is replaced with the description of each persona shown in Table \ref{tab:persona-descriptions}.

\paragraph{Implementation}
We describe the implementation for collecting LLM humor-preference data using persona prompting.
We used the three state-of-the-art LLMs, Gemini-3-Pro, GPT-5.1, and Claude-Sonnet-4.5, via their APIs.
For each API call, we set temperature to 1.0 and used default values for the other parameters.
To account for variability in API outputs, we collected three responses for each persona prompt.
In our analysis, we treated each response as an independent vote and used them to estimate BTL scores.

\subsection{Effectiveness and Limitations of Persona Prompting\label{appendix:persona_prompting_limitations}}
This subsection provides additional observations from our persona prompting experiments (\S\ref{section:analyzing_persona_prompting_efffect_on_LLM_preferences}) to clarify how persona prompts affect LLM-to-cluster alignment.
We focus on whether persona prompting changes the direction vs.\ magnitude of correlations and on heterogeneity across models, personas, and clusters.

First, in our experiments the different persona profiles tended to maintain the same direction of correlation with each user cluster (e.g., positive with C0, C2, C3 and negative with others); we did not observe a flip from positive to negative correlation when changing the persona.
Thus, persona prompting mainly modulates the magnitude of alignment rather than fully steering the model toward a different cluster, and it is insufficient on its own for full personalization.
Second, effects vary by model, persona, and cluster (e.g., stronger for Gemini than for GPT-5.1 or Claude; see Appendix~\ref{appendix:full_btl_scores_for_llm_analysis}).
Future studies should explore methods beyond persona prompting to align LLM humor preferences with individual users more precisely.

\section{Self-Consistency of LLM Funniest-Response Selections\label{appendix:llm_self_consistency}}
To assess how often the LLM selects the same response across the three API calls per prompt, we computed the proportion of prompts where all three calls agree (3of3), where at least two agree (2of3), and the average pairwise agreement (Pairwise) over 897 prompts (no\_persona setting; average 16 responses per prompt).
Table~\ref{tab:appendix-llm-self-consistency} reports these metrics for each model along with a random baseline (selecting one of 16 responses uniformly).
Agreement is substantially higher than the random baseline, indicating that the LLM's choices are not purely random and exhibit moderate self-consistency.

\begin{table}[t]
\centering
\small
\begin{tabular}{l r r r}
\toprule
\textbf{Model} & \textbf{3of3} & \textbf{2of3} & \textbf{Pairwise} \\
\midrule
Claude-Sonnet-4.5  & 0.313 & 0.499 & 0.480 \\
Gemini-3-Pro       & 0.202 & 0.462 & 0.356 \\
GPT-5.1            & 0.167 & 0.448 & 0.317 \\
Random baseline    & 0.004 & 0.176 & 0.063 \\
\bottomrule
\end{tabular}
\caption{Self-consistency of the three API calls per prompt (no\_persona): proportion of prompts where all three calls agree (3of3), at least two agree (2of3), and average pairwise agreement (Pairwise). Random baseline: expected value when selecting one of 16 responses uniformly at random.}
\label{tab:appendix-llm-self-consistency}
\end{table}

\section{Results of User-Cluster Preference Analysis\label{appendix:full_btl_scores_for_user_cluster_analysis}}
\begin{figure*}[t]
  \centering
  \includegraphics[width=\linewidth]{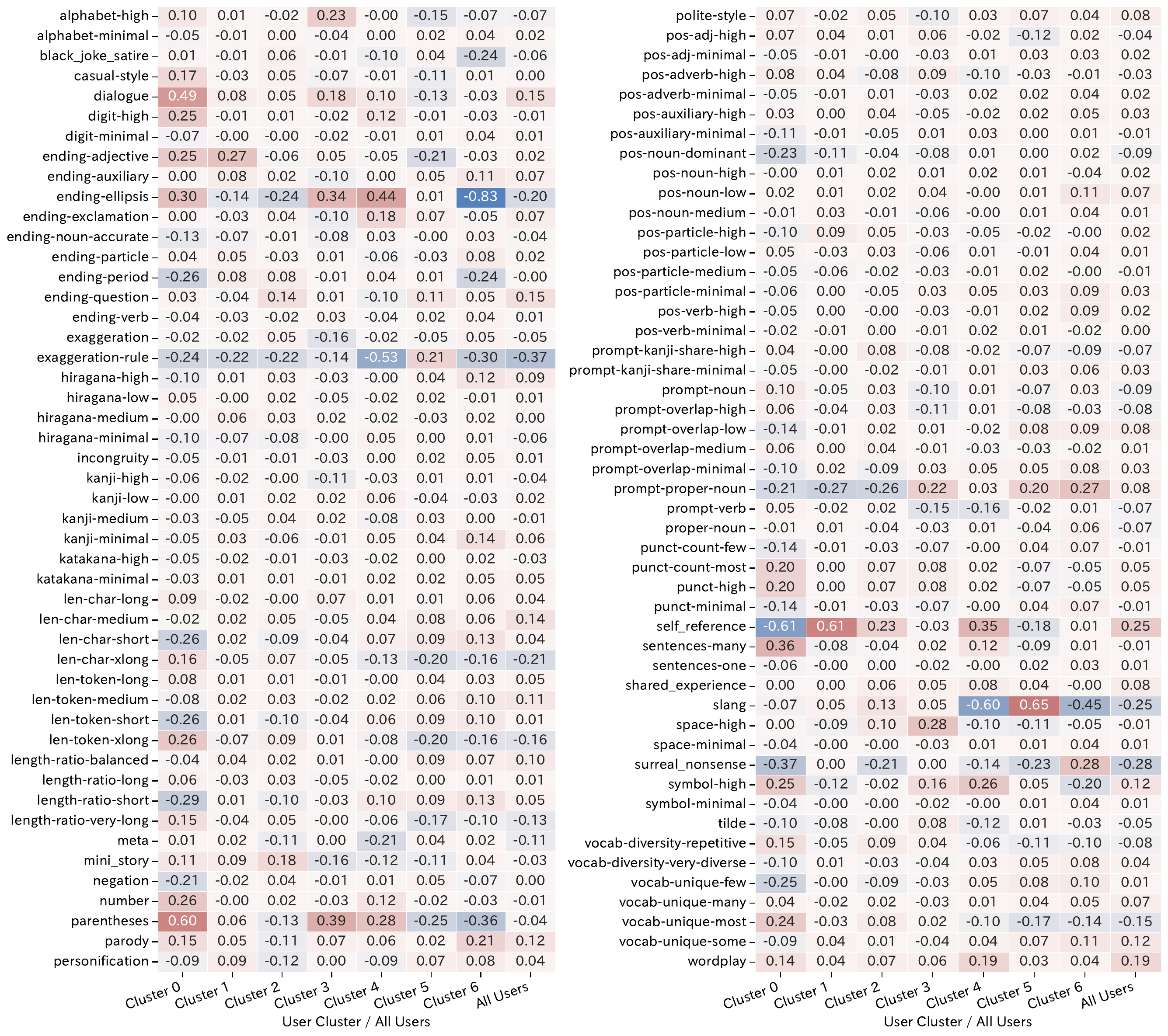}
  \caption{BTL scores of humor preference factors for each user cluster. ``All Users'' indicates the BTL scores calculated using all users without clustering.}
  \label{fig:btl-score-heatmap-for-user-cluster-all-users}
\end{figure*}

This section presents the results of the BTL scores of humor preference factors for each user cluster and all users.
Figure \ref{fig:btl-score-heatmap-for-user-cluster-all-users} shows the BTL scores of humor preference factors.
In this heatmap, rows represent humor preference factors, columns represent each user cluster and all users, and the color of each cell indicates the strength of the BTL score.

From the heatmap, we can visually confirm that the strengths of humor preference factors differ across user clusters.
For example, we observed that \texttt{ending-ellipsis} (responses ending with an ellipsis) has high BTL scores in clusters 0, 3, and 4 (0.30,  0.34, and 0.44, respectively), while showing low BTL scores in clusters 2 and 6 (-0.24 and -0.83, respectively), indicating different trends across clusters.

\section{Results of LLMs' Preference Analysis\label{appendix:full_btl_scores_for_llm_analysis}}
In this section, we discuss the trends in LLMs' humor preferences (\S\ref{appendix:llm_humor_preference_trends}), the effects of persona prompting (\S\ref{appendix:effect_of_persona_prompting}), the similarity between LLMs and user clusters in humor preferences (\S\ref{appendix:persona_llm_vs_user_cluster_humor_preference_correlation}), and the differences in humor preferences among persona LLMs (\S\ref{appendix:persona_llm_humor_preference_differences}).

\subsection{Factors that Influence LLMs' Humor Preferences\label{appendix:llm_humor_preference_trends}}
Figures \ref{fig:btl-score-heatmap-for-claude}, \ref{fig:btl-score-heatmap-for-gemini}, and \ref{fig:btl-score-heatmap-for-gpt-5.1} show the BTL scores of humor preference factors for Claude-Sonnet-4.5, Gemini-3-Pro, and GPT-5.1, respectively.
In each heatmap, rows represent humor preference factors, columns represent each persona, and the color of each cell indicates the strength of the BTL score.

We observed the following trends regarding humor preference factors favored or disfavored by LLMs:
Across all three models, LLMs consistently exhibit high BTL scores for overly long responses (\texttt{len-char-xlong}), the use of slang (\texttt{slang}), and dialogue punctuation (\texttt{dialogue}), while showing low BTL scores for responses of moderate length (\texttt{len-char-medium}) and responses with low vocabulary diversity (\texttt{vocab-unique-some}).
Additionally, we observed that the strength of BTL scores for certain factors, such as self-deprecation (\texttt{self-reference}), varied across models.

\subsection{Effect of Persona Prompting\label{appendix:effect_of_persona_prompting}}
\begin{figure*}[t]
  \centering
  \includegraphics[width=\linewidth]{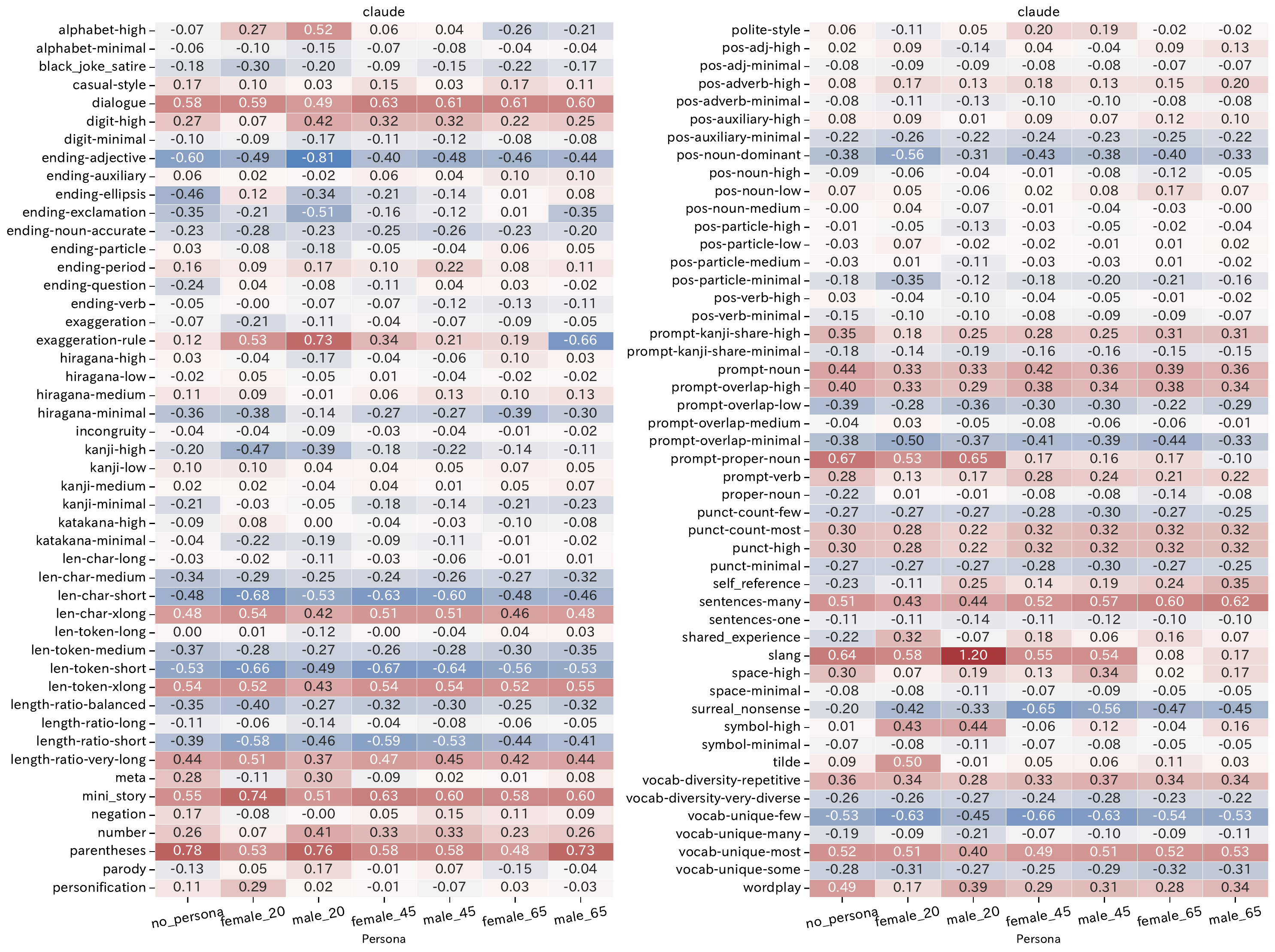}
  \caption{BTL scores of humor preference factors for Claude-Sonnet-4.5.}
  \label{fig:btl-score-heatmap-for-claude}
\end{figure*}

\begin{figure*}[t]
  \centering
  \includegraphics[width=\linewidth]{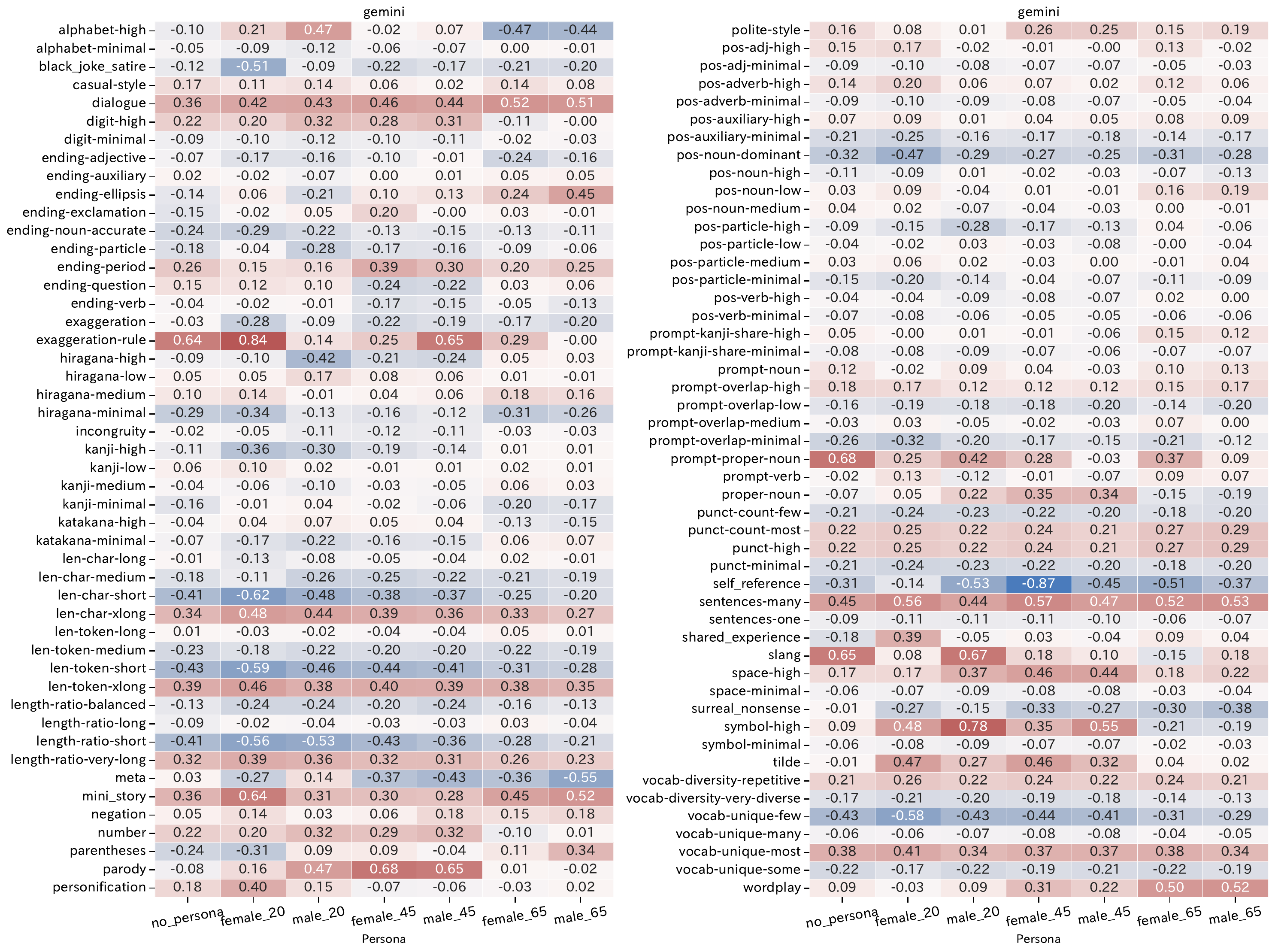}
  \caption{BTL scores of humor preference factors for Gemini-3-Pro.}
  \label{fig:btl-score-heatmap-for-gemini}
\end{figure*}

\begin{figure*}[t]
  \centering
  \includegraphics[width=\linewidth]{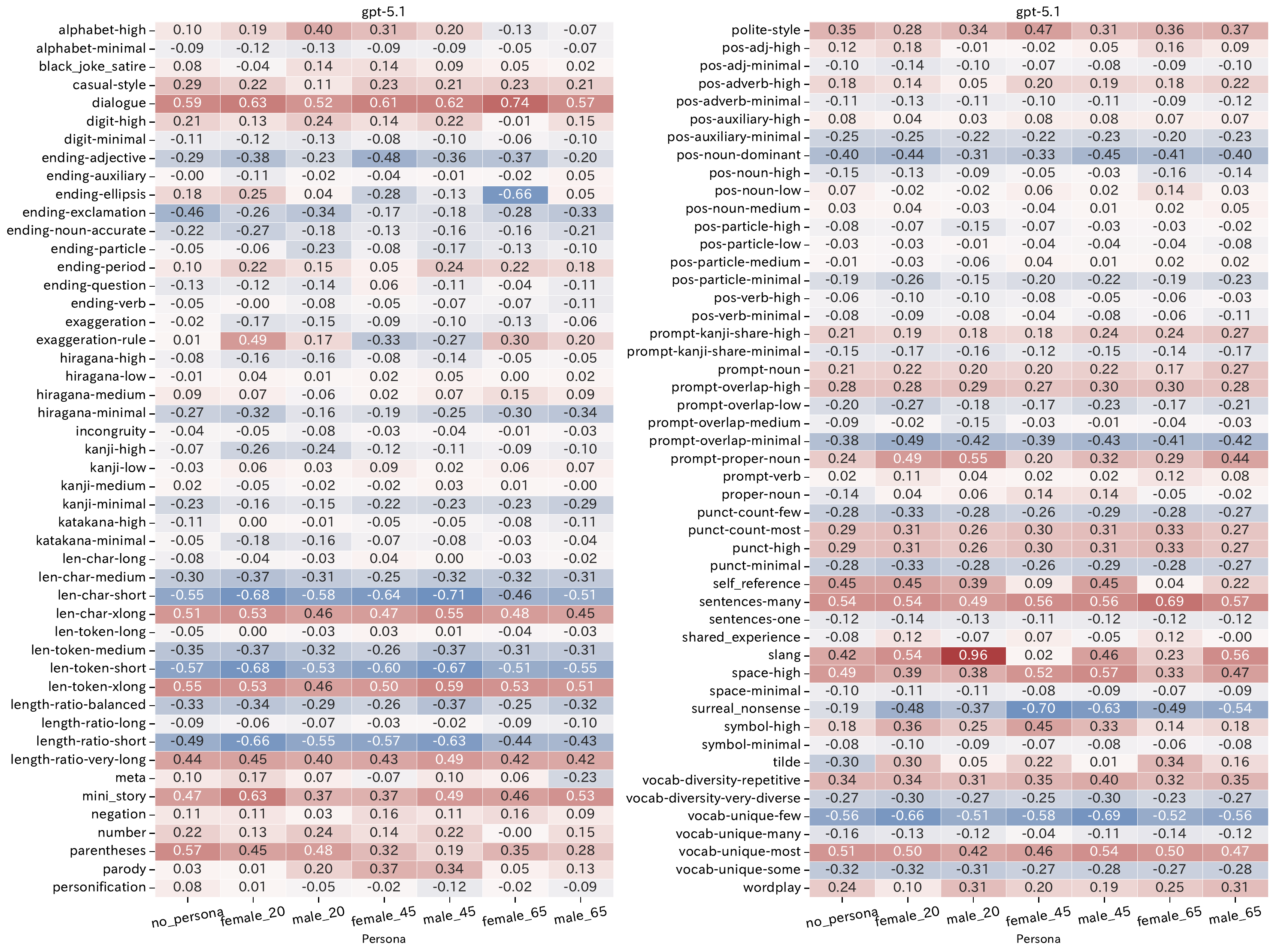}
  \caption{BTL scores of humor preference factors for GPT-5.1.}
  \label{fig:btl-score-heatmap-for-gpt-5.1}
\end{figure*}

From the heatmaps (Figure \ref{fig:btl-score-heatmap-for-claude}, \ref{fig:btl-score-heatmap-for-gemini}, \ref{fig:btl-score-heatmap-for-gpt-5.1}), we can visually confirm that the strengths of humor preference factors change with persona prompting compared to the no-persona setting (\texttt{no\_persona}).
For example, in Claude-Sonnet-4.5 (Figure \ref{fig:btl-score-heatmap-for-claude}), the BTL score for \texttt{exaggeration-rule} (responses containing exaggeration phrases) increases from 0.12 in the \texttt{no\_persona} setting to 0.73 in the \texttt{male\_20} persona, indicating a trend of changing humor preference factor strengths with persona prompting.

Additionally, the heatmaps show that while each persona of the models generally exhibits similar strengths of humor preference factors, some factors display different trends across personas.
For instance, in Claude-Sonnet-4.5 (Figure \ref{fig:btl-score-heatmap-for-claude}), \texttt{alphabet-high} (responses with a high alphabet ratio) shows high BTL scores in the \texttt{female\_20} and \texttt{male\_20} personas (0.27 and 0.52, respectively), while showing low BTL scores in the \texttt{female\_65} and \texttt{male\_65} personas (-0.26 and -0.21, respectively), indicating different trends across personas.

\subsection{Humor Preference Alignment between Persona LLM and User Cluster\label{appendix:persona_llm_vs_user_cluster_humor_preference_correlation}}

\begin{figure*}[t]
  \centering
  \begin{minipage}[t]{0.49\textwidth}
    \centering
    \includegraphics[width=\linewidth]{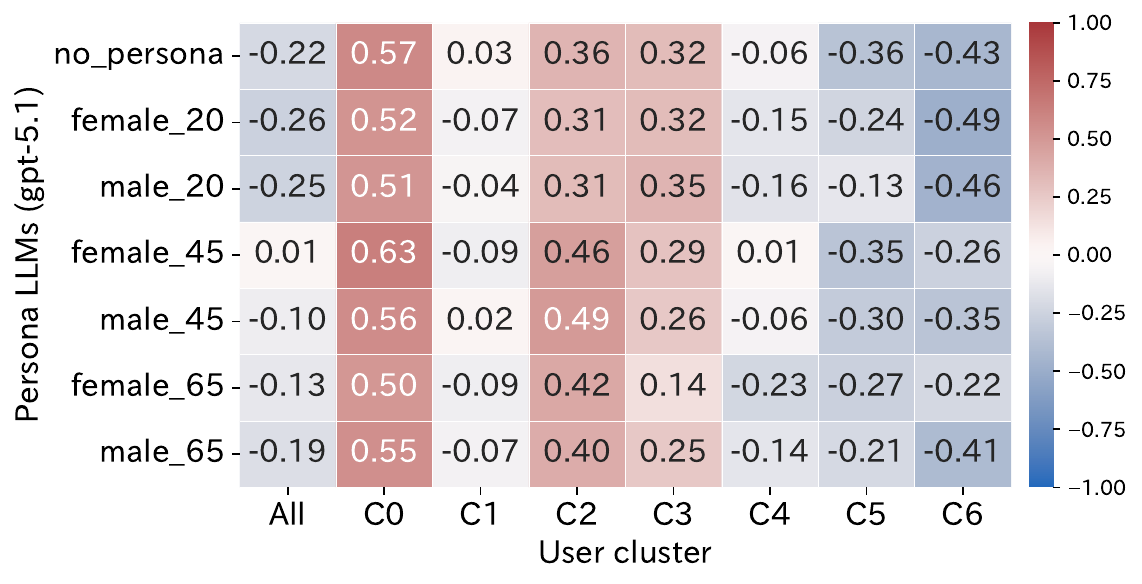}

    (a) GPT-5.1
  \end{minipage}
  \hfill
  \begin{minipage}[t]{0.49\textwidth}
    \centering
    \includegraphics[width=\linewidth]{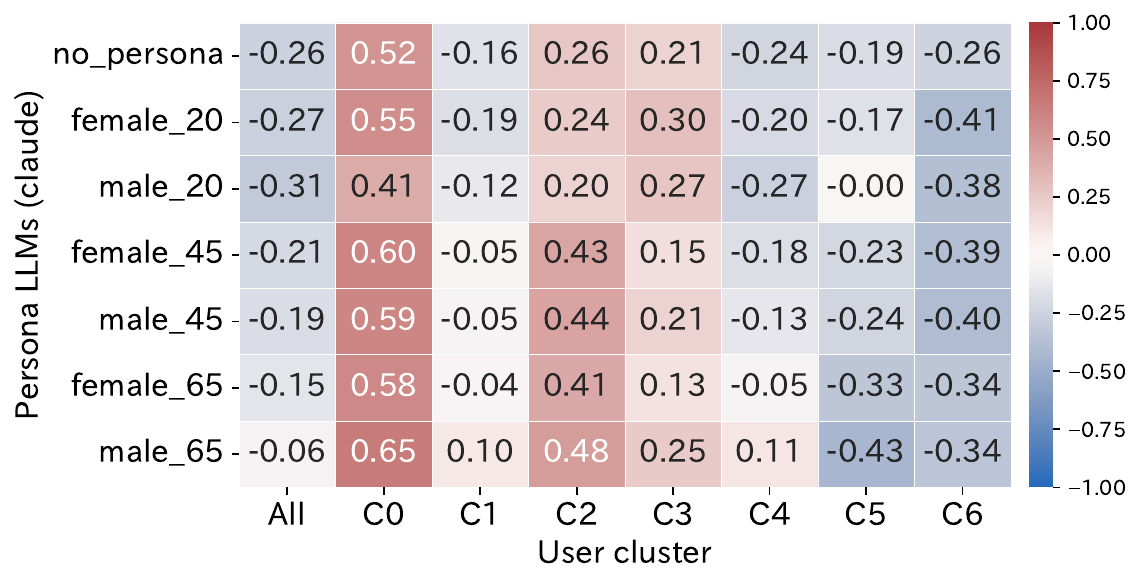}

    (b) Claude Sonnet 4.5
  \end{minipage}
  \caption{Pearson's correlation matrices between persona LLMs and user clusters, computed over the BTL scores of humor preference factors.}
  \label{fig:persona-vs-user-cluster-btl-pearson}
\end{figure*}

To address RQ3, we examined whether providing personas can influence an LLM’s humor preferences and thereby align the model with specific user clusters.
In this analysis, for each persona LLM, we computed the Pearson correlation coefficient between its BTL scores over preference factors and those of each user cluster. 
This enables a quantitative assessment of the similarity in humor preferences between persona LLMs and user clusters.
Figure \ref{fig:persona-vs-user-cluster-btl-pearson} presents heatmaps of the Pearson correlations for Claude-Sonnet-4.5 and GPT-5.1. Results for Gemini-3-Pro are already reported in the main text (Figure \ref{fig:gemini-persona-vs-user-cluster-btl-pearson}).
Each row corresponds to a persona LLM, each column corresponds to a user cluster, and each cell shows the Pearson correlation coefficient between the BTL scores of the persona and the user cluster.

The heatmaps indicate that changing the persona alters the correlation with each user cluster, suggesting that the model’s humor preference alignment varies across personas.
For example, for GPT-5.1, the correlation with C2 increases from 0.36 under the \texttt{no\_persona} setting to 0.49 under \texttt{male\_45}. Similar persona-dependent shifts are observed for Claude-Sonnet-4.5 and Gemini-3-Pro, indicating that providing personas can align LLMs with the humor preferences of particular clusters.

However, we also observed cases where the correlation coefficient remains largely unchanged regardless of the persona.
For instance, for GPT-5.1, the correlation with C1 ranges only from -0.09 to 0.03 across personas, showing no notable variation.
This suggests that persona prompting may be insufficient to align an LLM’s humor preferences for certain user clusters.
In the context of personalized humor evaluation and generation aimed at aligning with diverse user preferences, exploring alignment methods beyond persona prompting is an important direction for future work to better optimize for individual preferences.

\subsection{Humor Preference Differences between Persona LLMs\label{appendix:persona_llm_humor_preference_differences}}

\begin{figure*}[t]
  \centering
  \begin{minipage}[t]{0.32\textwidth}
    \centering
    \includegraphics[width=\linewidth]{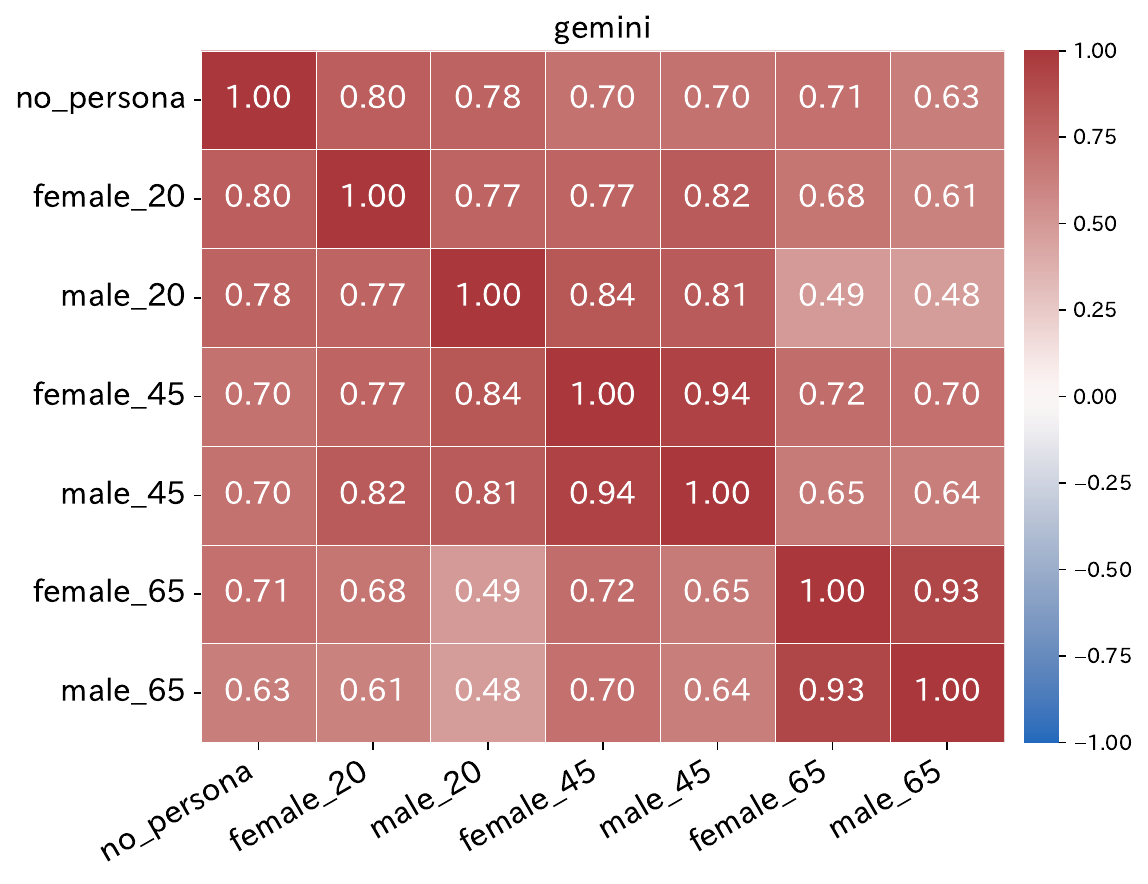}

    (a) Gemini 3 Pro
  \end{minipage}
  \hfill
  \begin{minipage}[t]{0.32\textwidth}
    \centering
    \includegraphics[width=\linewidth]{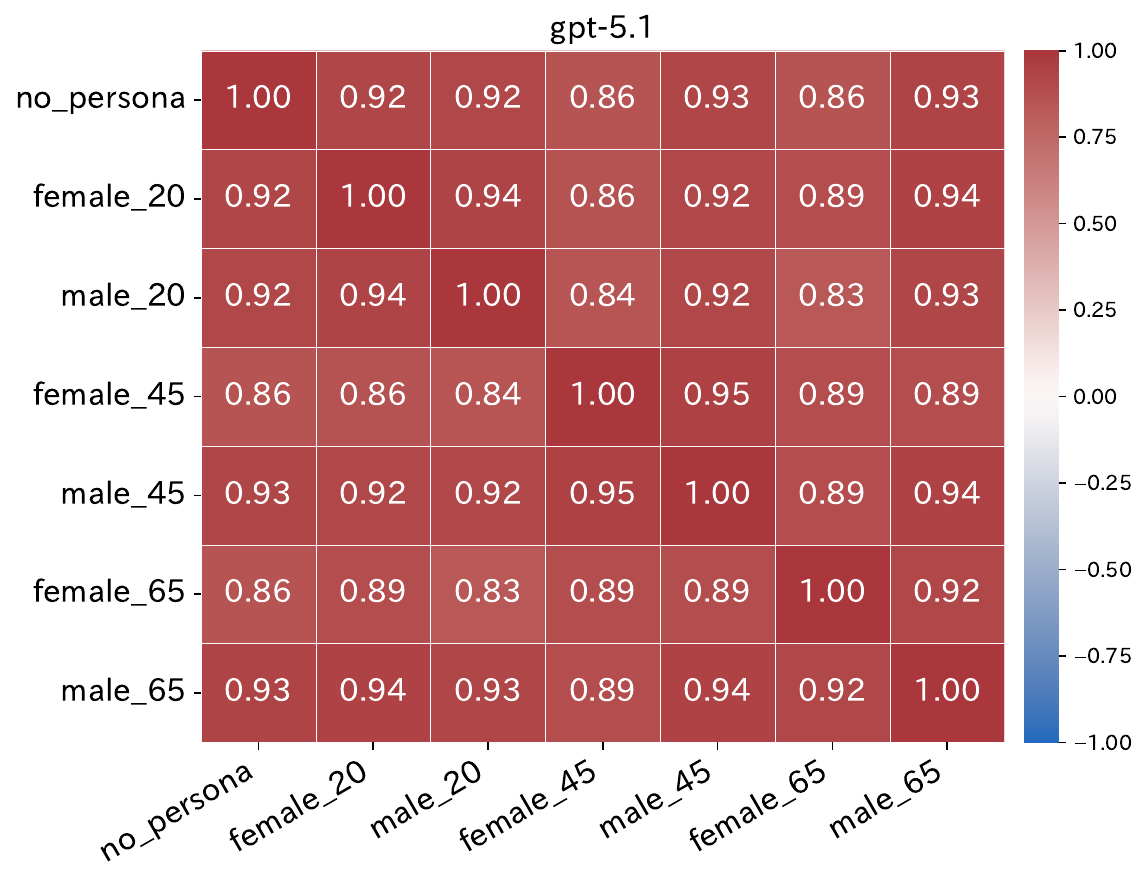}

    (b) GPT-5.1
  \end{minipage}
  \hfill
  \begin{minipage}[t]{0.32\textwidth}
    \centering
    \includegraphics[width=\linewidth]{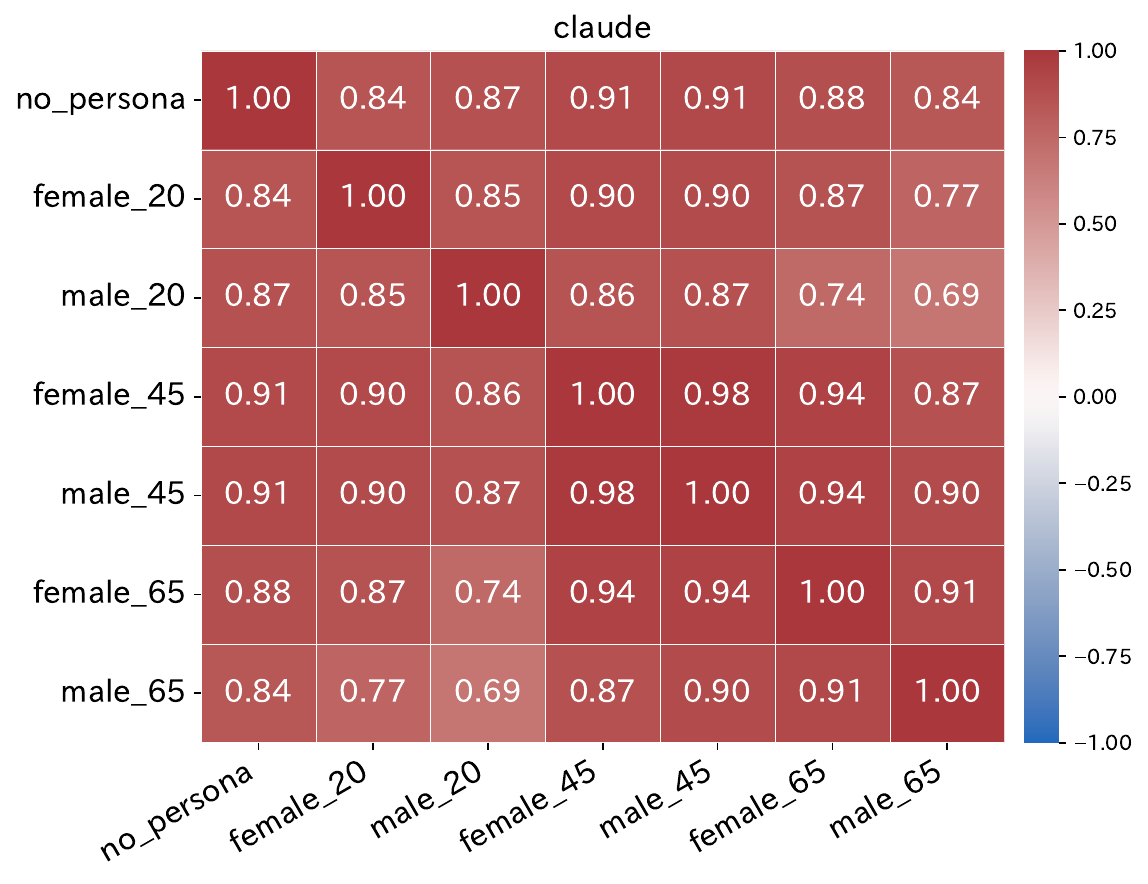}

    (c) Claude Sonnet 4.5
  \end{minipage}
  \caption{Pearson's correlation matrices between persona LLMs, computed over the BTL scores of humor preference factors.}
  \label{fig:persona-to-persona-btl-pearson}
\end{figure*}

Finally, we quantitatively analyzed whether assigning different personas to an LLM induces differences in humor preferences across persona LLMs.
Specifically, we computed the Pearson correlation coefficients between the BTL scores of humor-preference factors across persona LLMs.
This enables a quantitative assessment of whether humor preferences are similar across persona LLMs.
For example, if the correlation coefficient between different personas (e.g., \texttt{male\_20} and \texttt{female\_65}) is low, this indicates that the tendencies of humor preference differ across personas; conversely, a high coefficient suggests similar humor preferences.
Figure~\ref{fig:persona-to-persona-btl-pearson} shows the Pearson correlation coefficients of BTL scores across personas for Gemini-3-Pro, GPT-5.1, and Claude-Sonnet-4.5.
Each row and column corresponds to a persona, and each cell represents the Pearson correlation coefficient between the BTL scores of the corresponding persona pair.

From this heatmap, we can visually confirm that the correlations between personas tend to be high on average.
This suggests that even when different personas are provided, the LLM is likely to exhibit similar humor preferences.
One plausible explanation is overlap in pretraining data and similar post-training procedures; however, we did not test this directly.
Future work should test this hypothesis by comparing these API-only models with open-weight models and models trained on different corpora, examining whether shared training data contributes to the observed LLM-to-LLM similarity relative to the user clusters in our setting.

Another noteworthy point is that the overall pattern of inter-persona correlations differs substantially across models.
Gemini-3-Pro tends to exhibit larger variations in humor-preference correlations across different personas, whereas GPT-5.1 and Claude-Sonnet-4.5 show consistently high correlations across most persona pairs.
For instance, the correlation in humor preference between \texttt{female\_65} and \texttt{male\_20} is 0.49 (a moderate positive correlation) for Gemini-3-Pro, but 0.83 and 0.74 (strong positive correlations) for GPT-5.1 and Claude-Sonnet-4.5, respectively.
These results suggest that Gemini-3-Pro is more influenced by persona-induced changes in humor preferences, while the other two models are less affected.

One possible explanation is the impact of reasoning effort.
Gemini-3-Pro performs extensive reasoning by default in the API, whereas the other models do not.
Introducing reasoning effort may make the LLM more likely to exhibit persona-specific behaviors, which could in turn change humor preferences.
In future work, we plan to systematically investigate the effects of varying the scale of reasoning effort and the choice of personas. Such analyses are expected to provide more concrete evidence for the usefulness of persona prompting aimed at alignment to specific user clusters.
\end{document}